\documentclass[usenames,dvipsnames, 11pt,a4paper]{article}

\usepackage{times,latexsym}
\usepackage{url}
\usepackage[T1]{fontenc}
\usepackage{authblk} % Added by authors
\makeatletter
\newcommand\email[2][]%
   {\newaffiltrue\let\AB@blk@and\AB@pand
      \if\relax#1\relax\def\AB@note{\AB@thenote}\else\def\AB@note{\relax}%
        \setcounter{Maxaffil}{0}\fi
      \begingroup
        \let\protect\@unexpandable@protect
        \def\thanks{\protect\thanks}\def\footnote{\protect\footnote}%
        \@temptokena=\expandafter{\AB@authors}%
        {\def\\{\protect\\\protect\Affilfont}\xdef\AB@temp{#2}}%
         \xdef\AB@authors{\the\@temptokena\AB@las\AB@au@str
         \protect\\[\affilsep]\protect\Affilfont\AB@temp}%
         \gdef\AB@las{}\gdef\AB@au@str{}%
        {\def\\{, \ignorespaces}\xdef\AB@temp{#2}}%
        \@temptokena=\expandafter{\AB@affillist}%
        \xdef\AB@affillist{\the\@temptokena \AB@affilsep
          \AB@affilnote{}\protect\Affilfont\AB@temp}%
      \endgroup
       \let\AB@affilsep\AB@affilsepx
}
\makeatother

\usepackage[acceptedWithA]{tacl2018v2}

\usepackage{amsfonts}
\usepackage{amsmath}
\usepackage{amssymb}
\usepackage[shortlabels]{enumitem}
\usepackage[font={small}]{caption}
\usepackage[multiple]{footmisc}
\usepackage{graphicx}
\usepackage{multirow}
\usepackage{stmaryrd}
\usepackage{algorithm}
\usepackage[noend]{algpseudocode}

\usepackage{xspace,mfirstuc,tabulary}

\newif\ifcomments
\commentstrue 
\ifcomments
    \newcommand\mg[1]{\textcolor{purple}{[MG: #1]}}
    \newcommand\tw[1]{\textcolor{red}{[TW: #1]}}
    \newcommand\jb[1]{\textcolor{blue}{[JB: #1]}}
    \newcommand\ag[1]{\textcolor{green}{[AG: #1]}}
    
\else
    \providecommand{\mg}[1]{}
    \providecommand{\tw}[1]{}
    \providecommand{\jb}[1]{}
    \providecommand{\ag}[1]{}
\fi

\newcommand\copybase{\textsc{Copy}}
\newcommand\rulebased{\textsc{RuleBased}}
\newcommand\seqtoseq{\textsc{Seq2Seq}}
\newcommand\dynamic{\textsc{S2SDynamic}}
\newcommand\copynet{\textsc{Copynet}}
\newcommand\datasetname{\textsc{Break}}
\newcommand\comment[1]{}
\newcommand\decomplist{\textbf{s}}
\newif\iftaclinstructions
\taclinstructionsfalse % AUTHORS: do NOT set this to true
\iftaclinstructions
\renewcommand{\confidential}{}
\renewcommand{\anonsubtext}{(No author info supplied here, for consistency with TACL-submission anonymization requirements)}
\newcommand{\instr}
\fi

\iftaclpubformat % this "if" is set by the choice of options

\else

\fi

\title{\textsc{Break} It Down: A Question Understanding Benchmark}

\author[1,3]{\bf Tomer Wolfson}
\author[1,3]{\bf Mor Geva}
\author[1]{\bf Ankit Gupta}
\author[3]{\\\bf Matt Gardner}
\author[2,3]{\bf Yoav Goldberg}
\author[1]{\bf Daniel Deutch}
\author[1,3]{\bf Jonathan Berant}

{
\makeatletter
\renewcommand\AB@affilsepx{\quad \protect\Affilfont}
\makeatother
\affil[1]{Tel Aviv University}
\affil[2]{Bar-Ilan University}
\affil[3]{Allen Institute for AI}}

\email{}

\email{\normalsize	 \texttt{$\{$tomerwol,morgeva$\}$@mail.tau.ac.il}, \quad \texttt{ankitgupta.iitkanpur@gmail.com},}
\email{\normalsize \texttt{mattg@allenai.org}, \quad \texttt{yoav.goldberg@gmail.com},}
\email{\normalsize \texttt{danielde@post.tau.ac.il},\quad \texttt{joberant@cs.tau.ac.il}}

\date{}

\begin{document}
\maketitle

\begin{abstract}
%A core part of question understanding is the ability to break down a question into the sequence of steps that need to be taken to compute its answer.
  Understanding natural language questions entails the ability to break down a question into the requisite steps for computing its answer. In this work, we introduce a Question Decomposition Meaning Representation (QDMR) for questions. QDMR constitutes the ordered list of steps, expressed through natural language, that are necessary for answering a question. We develop a crowdsourcing pipeline, showing that quality QDMRs can be annotated at scale, and release the \datasetname{} dataset, containing over 83K pairs of questions and their QDMRs.
  We demonstrate the utility of QDMR by showing that (a) it can be used to improve open-domain question answering on the \textsc{HotpotQA} dataset, (b) it can be deterministically converted to a pseudo-SQL formal language, which can alleviate annotation in semantic parsing applications.
  Last, we use \datasetname{} to train a sequence-to-sequence model with copying that parses questions into QDMR structures, and show that it substantially outperforms several natural baselines.
  
  %and observe that while reasonable decompositions can be generated,
  %they fall short compared to human-produced QDMRs.
  % JB: SARI is not that known let's not focus on that metric.
  %By applying an encoder-decoder model on this dataset we show that the best systems only achieve XX SARI score, while expert human performance nears XX.
  %JB: the utility of QDMRs should be in the abstract 
  %Last, to demonstrate the practical utility of QDMR, we show that (a) it can be
  %directly used to improve performance on the \textsc{HotpotQA} reading comprehension
  %dataset, (b) it can be deterministically converted to a pseudo-SQL formal
  %language.
\end{abstract}

\section{Introduction}

\begin{figure}[t]
  \includegraphics[trim={0.2cm 0cm 18.4cm 1.9cm}, clip, width=8cm,height=8.8cm]{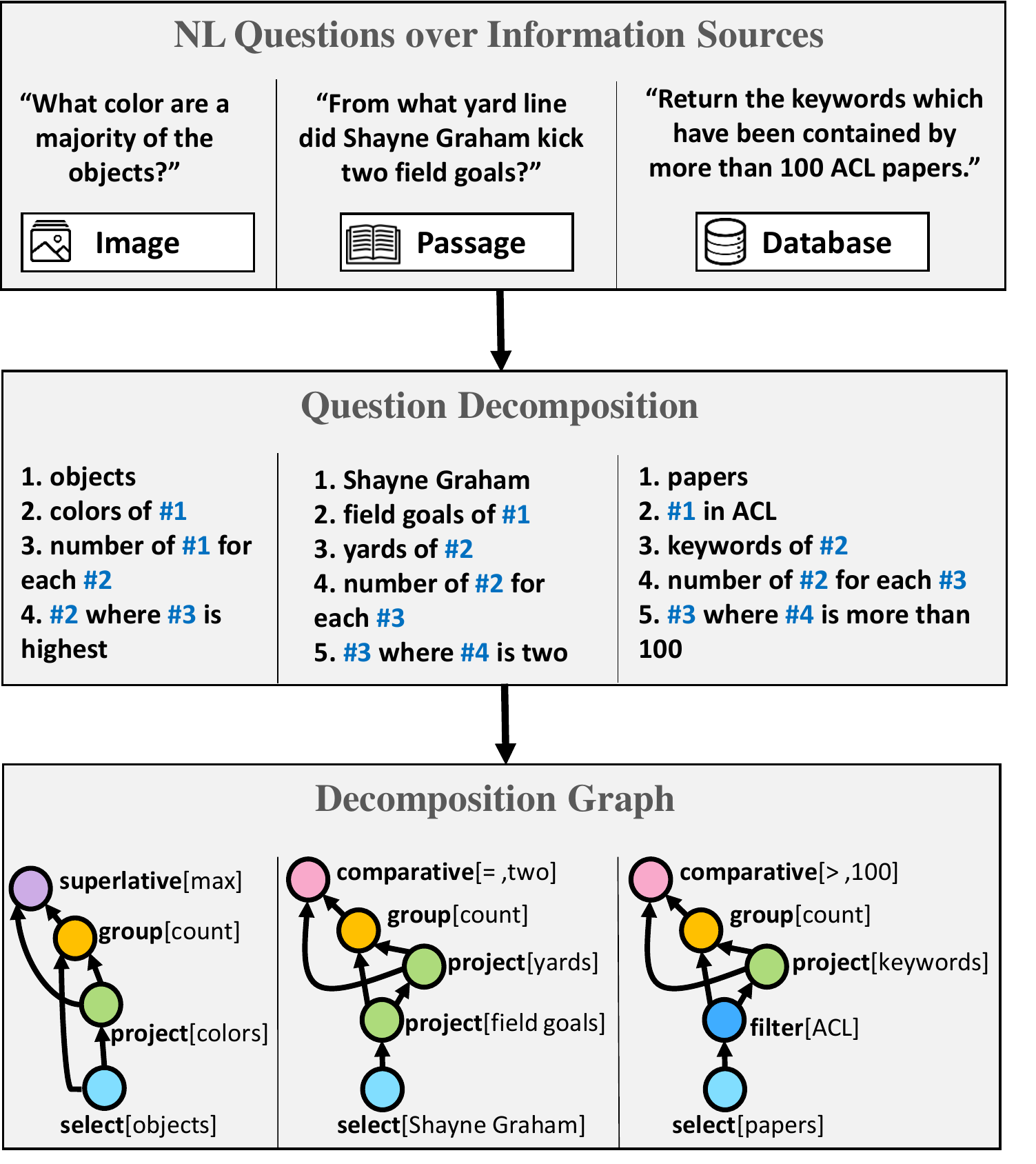}
  \caption{Questions over different sources share a similar compositional structure. Natural language questions from multiple sources (top) are annotated with the QDMR formalism (middle) and deterministically mapped into a pseudo-formal language (bottom).}
\label{figure:motivation}
\end{figure}

%JB: reasoning is important now and complex QA is the vehicle to test it.
Recently, increasing work has been devoted to models that can reason and integrate information from multiple parts of an input. This includes reasoning over images \cite{antol2015vqa, johnson2017clevr, suhr2018corpus, Hudson_2019_CVPR}, paragraphs \cite{dua2019drop}, documents \cite{welbl2017constructing, talmor2018web,yang2018HotpotQAAD}, tables
\cite{pasupat2015compositional} and more. \emph{Question answering} (QA) is commonly used to test  the ability to reason, where a complex natural language question is posed, and is to be answered given a particular context (text, image, etc.). 
%JB: Reasons why it makes sense to have question understanding models.
Although questions often share structure across tasks and modalities, understanding the language of complex questions has thus far been dealt within each task in isolation. Consider the questions in Figure \ref{figure:motivation}, all of which express operations such as fact chaining and counting. Additionally, humans can take a complex question and break it down into a sequence of simpler questions even when they are unaware of what or where the answer is. 
This ability, to compose and decompose questions, lies at the heart of human language \cite{Pelletier1994} and allows us to tackle previously unseen problems.
Thus, better question understanding models should improve performance and generalization in tasks that require multi-step reasoning or that do not have access to substantial amounts of data.

In this work we propose \emph{question understanding} as a standalone language understanding task. We introduce a formalism for representing the meaning of questions that relies on \emph{question decomposition}, and is agnostic to the information source.
Our formalism, Question Decomposition Meaning Representation (QDMR), is inspired by database query languages (SQL; SPARQL), and by semantic parsing \cite{zelle96geoquery, zettlemoyer05ccg, clarke10world}, in which questions are given full meaning representations.

We express complex questions via simple (``atomic'') questions that can be executed in sequence to answer the original question.
Each atomic question can be mapped into a small set of \textit{formal operations}, where each operation either selects a set of entities, retrieves information about their attributes, or aggregates information over entities.
While this has been formalized in knowledge-base (KB) query languages \cite{chamberlin1974sequel}, the same intuition can be applied to other modalities, such as images and text. 
QDMR abstracts away the context needed to answer the question, allowing in principle to query multiple sources for the same question.
%Our formalism, Question Decomposition Meaning Representation (QDMR), represents the meaning of questions in a manner that is agnostic to the information source. Inherently expressed in natural language, QDMR is inspired by relational algebra operators \cite{codd1970relational} and by semantic parsing \cite{zelle96geoquery, zettlemoyer05ccg, clarke10world, liang13cl}, where a question is translated into a full meaning representation.

In contrast to semantic parsing, QDMR operations are expressed through \emph{natural language}, facilitating annotation at scale by non-experts.
Figure \ref{figure:motivation} presents examples of complex questions on three different modalities. The middle box lists the natural language decompositions provided for each question, and the bottom box displays their corresponding formal queries. 

QDMR serves as the formalism for creating \datasetname{}, a question
decomposition dataset of 83,978 questions over ten datasets and three modalities.
\datasetname{} is collected via crowdsourcing, with a user interface that allows us to train crowd workers to produce quality decompositions (\S\ref{sec:data_collection}). Validating the quality of annotated structures reveals 97.4\% to be correct (\S \ref{sec:data_analysis}).

We demonstrate the utility of QDMR in two setups.
First, we regard the task of open-domain QA over multi-hop questions from the \textsc{HotpotQA} dataset. Combining QDMR structures in \datasetname{} with an RC model \cite{min2019multi} improves F$_1$ from 43.3 to 52.4 (\S\ref{sec:application_qa}).
Second, we show that decompositions in \datasetname{} possess high annotation consistency, which indicates that annotators produce high-quality QDMRs (\S\ref{subsec:annotation_consistency}). In \S\ref{subsec:application_low} we discuss how these QDMRs can be used as a strong proxy for full logical forms in semantic parsing. 

% TW - changed following the revision
%%%Second, we show that 99.2\% of the QDMRs in \datasetname{} can be converted using a deterministic algorithm into corresponding pseudo-SQL formulas (\S\ref{subsec:application_low}), out of which 93\% are fully decomposed. This indicates that annotators produce high-quality and consistent annotations which can be used as a strong proxy for full logical forms in semantic parsing. 

We use \datasetname{} to train a neural QDMR parser that maps questions into QDMR representations, based on a sequence-to-sequence model with copying \cite{gu2016copying}. Manual analysis of generated structures reveals an accuracy of 54\%, showing that automatic QDMR parsing is possible, though still far from human performance (\S\ref{sec:question_decomposition}).

To conclude, our contributions are:
\begin{itemize}[topsep=0pt, itemsep=0pt, leftmargin=.2in, parsep=0pt]
    \item Proposing the task of question understanding and introducing the QDMR formalism for representing the meaning of questions  (\S\ref{sec:decomposition_formalism})
    \item The \datasetname{} dataset, which consists of 83,978 examples sampled from 10 datasets over three distinct information sources (\S\ref{sec:data_collection})
    \item Showing how QDMR can be used to improve open-domain question answering (\S\ref{sec:application_qa}),  as well as alleviate the burden of annotating logical forms in semantic parsing (\S\ref{subsec:application_low})
    \item A QDMR parser based on a sequence-to-sequence model with copying mechanism (\S\ref{sec:question_decomposition})
\end{itemize}

The \datasetname{} dataset, models and entire codebase are publicly available at: \url{https://allenai.github.io/Break/}.

\section{Question Decomposition Formalism}
\label{sec:decomposition_formalism}

\begin{table*}[t!]
\begin{center}
\scriptsize
\begin{tabular}{p{1.5cm} p{4.1cm} p{4.1cm} p{4.7cm}}
\hline \bf Operator & \bf Template / Signature & \bf Question & \bf Decomposition  \\ \hline
{\color{NavyBlue} \bf Select} & Return [entities] \newline $\texttt{w} \rightarrow \texttt{S$_\texttt{e}$}$ & How many touchdowns were scored overall? &  1{\color{NavyBlue}\bf . Return touchdowns} \newline 2. Return the number of \#1 \\ \hline
{\color{Orange} \bf Filter} & Return [ref] [condition] \newline $\texttt{S$_\texttt{o}$}\texttt{,} \texttt{w} \rightarrow \texttt{S$_\texttt{o}$}$ & I would like a flight from Toronto to San Diego please. & 1. Return flights \newline {\color{Orange} \bf 2. Return \#1 from Toronto \newline 3. Return \#2 to San Diego} \\ \hline
{\color{CadetBlue} \bf Project} & Return [relation] of [ref] \newline $\texttt{w}\texttt{,} \texttt{S$_\texttt{e}$} \rightarrow \texttt{S$_\texttt{o}$}$ & Who is the head coach of the Los Angeles Lakers? & 1. Return the Los Angeles Lakers \newline {\color{CadetBlue} \bf 2. Return the head coach of \#1} \\ \hline
{\color{YellowOrange} \bf Aggregate} & Return [aggregate] of [ref] \newline $\texttt{w}_{\texttt{agg}}\texttt{,} \texttt{S$_\texttt{o}$} \rightarrow \texttt{n}$ & How many states border Colorado? & 1. Return Colorado  \newline 2. Return border states of \#1 \newline {\color{YellowOrange} \bf 3. Return the number of \#2} \\ \hline
{\color{BlueViolet} \bf Group} & Return [aggregate] [ref1] for each [ref2] \newline $\texttt{w}_{\texttt{agg}}\texttt{,} \texttt{S$_\texttt{o}$,} \texttt{S$_\texttt{e}$} \rightarrow \texttt{S$_\texttt{n}$}$ & How many female students are there in each club? & 1. Return clubs \newline
2. Return female students of \#1 \newline
{\color{BlueViolet} \bf 3. Return the number of \#2 for each \#1 } \\ \hline
{\color{Green} \bf Superlative} & Return [ref1] where [ref2] is [highest / lowest] \newline $\texttt{S$_\texttt{e}$,} \texttt{S$_\texttt{n}$,} \texttt{w}_{\texttt{sup}} \rightarrow \texttt{S$_\texttt{e}$}$ & What is the keyword, which has been contained by the most number of papers?  & 1. Return papers \newline 
2. Return keywords of \#1 \newline
3. Return the number of \#1 for each \#2 \newline
{\color{Green} \bf 4. Return \#2 where \#3 is highest} \\ \hline
{\color{Plum} \bf Comparative} & Return [ref1] where [ref2] [comparison] [number] \newline $\texttt{S$_\texttt{e}$,} \texttt{S$_\texttt{n}$,} \texttt{w}_{\texttt{com}}\texttt{,} \texttt{n} \rightarrow \texttt{S$_\texttt{e}$}$ & Who are the authors who have more than 500 papers? & 1. Return authors \newline
2. Return papers of \#1 \newline
3. Return the number of \#2 for each of \#1 \newline
{\color{Plum} \bf 4. Return \#1 where \#3 is more than 500} \\ \hline
{\color{Magenta} \bf Union} & Return [ref1] , [ref2] \newline $\texttt{S$_\texttt{o}$,} \texttt{S$_\texttt{o}$} \rightarrow \texttt{S$_\texttt{o}$}$ & Tell me who the president and vice-president are? & 1. Return the president \newline
2. Return the vice-president \newline
{\color{Magenta} \bf 3. Return \#1 , \#2}  \\ \hline
{\color{Bittersweet} \bf Intersection} & Return [relation] in both [ref1] and [ref2] \newline $\texttt{w,} \texttt{S$_\texttt{e}$,} \texttt{S$_\texttt{e}$} \rightarrow \texttt{S$_\texttt{o}$}$ & Show the parties that have representatives in both New York state and representatives in Pennsylvania state. & 1. Return representatives \newline 2. Return \#1 in New York state \newline 3. Return \#1 in Pennsylvania state \newline {\color{Bittersweet} \bf 4. Return parties in  both \#2 and \#3} \\ \hline
{\color{RawSienna} \bf Discard} & Return [ref1] besides [ref2] \newline $\texttt{S$_\texttt{o}$,} \texttt{S$_\texttt{o}$} \rightarrow \texttt{S$_\texttt{o}$}$ & Find the professors who are not playing Canoeing. & 1. Return professors \newline 2. Return \#1 playing Canoeing \newline {\color{RawSienna}  \bf 3. Return \#1 besides \#2} \\ \hline
{\color{RedViolet} \bf Sort} & Return [ref1] sorted by [ref2] \newline $\texttt{S$_\texttt{e}$,} \texttt{S$_\texttt{n}$} \rightarrow \texttt{$\langle \texttt{e}_1...\texttt{e}_k \rangle$}$ & Find all information about student addresses, and sort by monthly rental. & 1. Return students \newline 2. Return addresses of \#1 \newline 3. Return monthly rental of  \#2 \newline {\color{RedViolet} \bf 4. Return \#2 sorted by \#3}
 \\ \hline
{\color{OliveGreen} \bf Boolean} & Return [if / is] [ref1] [condition] [ref2] \newline $\texttt{S$_\texttt{o}$,} \texttt{w,} \texttt{S$_\texttt{o}$} \rightarrow \texttt{b}$ & Were Scott Derrickson and Ed Wood of the same nationality? & ... \newline 3. Return the nationality of \#1 \newline 4. Return the nationality of \#2 \newline {\color{OliveGreen} \bf 5. Return if \#3 is the same as \#4} \\ \hline
{\color{Cyan} \bf Arithmetic} & Return the [arithmetic] of [ref1] and [ref2] \newline $\texttt{w}_{\texttt{ari}}\texttt{,} \texttt{n,} \texttt{n} \rightarrow \texttt{n}$ & How many more red objects are there than blue objects? & ... \newline 3. Return the number of \#1 \newline 4. Return the number of \#2 \newline {\color{Cyan} \bf 5. Return the difference of \#3 and \#4} \\ \hline
\end{tabular}
\end{center}
\caption{\label{table:operators} The 13 operator types of QDMR steps. Listed are, the natural language template used to express the operator, the operator signature and an example question that uses the query operator in its decomposition.}
\end{table*}

% TW - inspiration for QDMR
In this section we define the QDMR formalism for domain agnostic question decomposition. 

QDMR is primarily inspired by SQL  \cite{codd1970relational, chamberlin1974sequel}.
However, while SQL was designed for relational databases, QDMR also aims to capture the meaning of questions over unstructured sources such as text and images.
Thus, our formalism abstracts away from SQL by assuming an underlying ``idealized'' KB, which contains all entities and relations expressed in the question. This abstraction enables QDMR to be unrestricted to a particular modality, with its operators to be executed also against text and images, while allowing in principle to query multiple modalities for the same question.\footnote{A system could potentially answer \textit{``Name the political parties of the most densely populated country''}, by retrieving \textit{``the most densely populated country''} using a database query, and \textit{``the political parties of \#1''} via an RC model.}

% TW - omitted for lack of space
%As it is based on natural language phrases, QDMR can be annotated by non-experts. Thus, it has much broader coverage to questions where the information is not in a structured KB, and can be annotated at a low cost at scale. We will show in \S\ref{subsec:application_low} that QDMR can be deterministically converted into a formal language resembling SQL.

\paragraph{QDMR Definition}\label{subsec:definition}
%We now define the QDMR formalism for natural language question decomposition.
Given a question $x$, its QDMR is a sequence of $n$
steps, $\decomplist = \langle s^1, ..., s^n \rangle$,
where each step $s^i$ corresponds to a single query operator $f^i$ (see Table~\ref{table:operators}).
A step, $s^i$ is a sequence of tokens, $s^i = (s^{i}_1,...,s^{i}_{m_i})$, where a token $s^{i}_k$ is either a word from a predefined lexicon $L_x$ (details in \S\ref{sec:data_collection}) or a \emph{reference token}, referring to the result of a previous step $s^j$, where $j<i$.
The last step, $s^n$ returns the answer to $x$.

\begin{figure}[t]
  \includegraphics[trim={0cm 0.16cm 14.15cm 11.8cm}, clip, width=7.5cm,height=2.6cm]{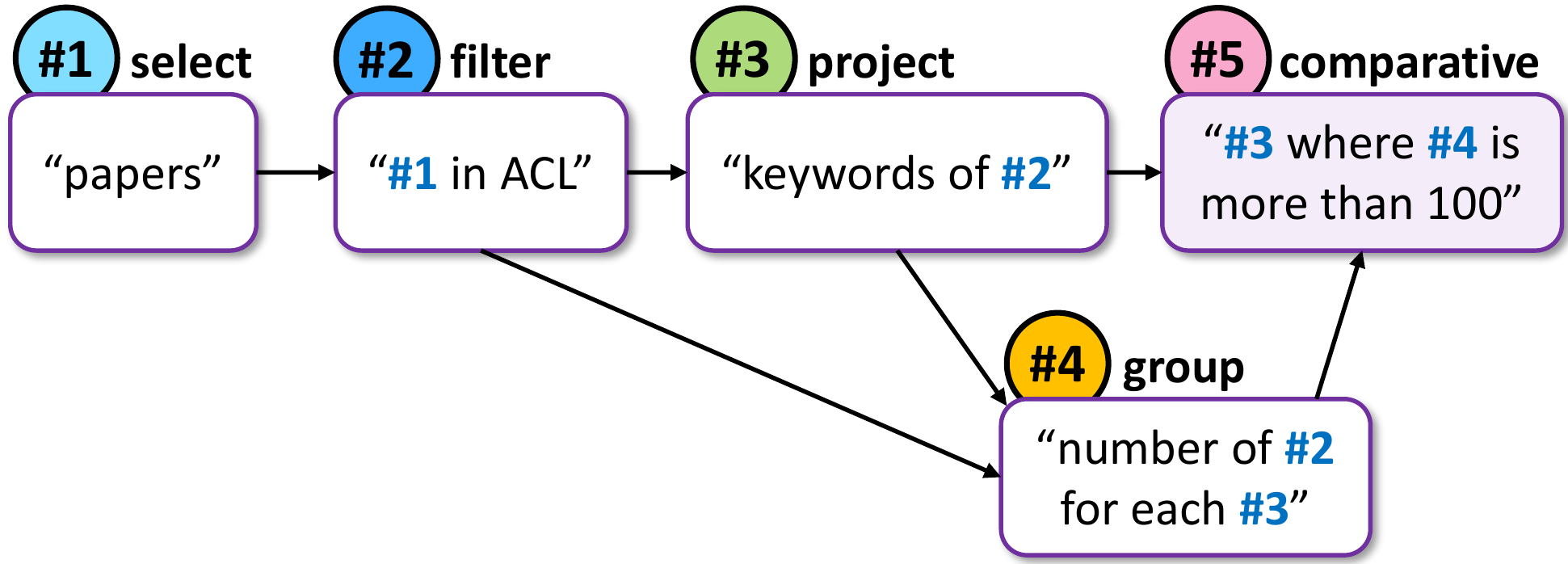}
\caption{QDMR of the question \textit{``Return the keywords which have been contained by more than 100 ACL papers.''}, represented as a decomposition graph.}
\label{fig_decomposition_dag}
\end{figure}

\paragraph{Decomposition Graph}\label{subsec:qdmr_dag}
QDMR structures can be represented as a directed acyclic graph (DAG), used for evaluating QDMR parsing models (\S\ref{sec:eval_metrics}). Given QDMR, $\decomplist = \langle s^1, ..., s^n \rangle$, each step $s^i$ is a node in the graph, labeled by its sequence of tokens and index $i$. Edges in the graph are induced by \emph{reference tokens} to previous steps. Node $s^i$ is connected by an incoming edge $(s^j, s^i)$, if $ref[s^j] \in (s^{i}_1,...,s^{i}_{m_i})$. That is, if one of the tokens in $s^i$ is a reference to $s_j$. Figure~\ref{fig_decomposition_dag} displays a sequence of QDMR steps, represented as a DAG.

\begin{table}[t]
\begin{center}
\tiny
\begin{tabular}{p{1.25cm}|p{5cm}}
\hline \bf Function & \bf Description   \\ \hline
$\mathrm{agg}$ & Given a phrase $w_{agg}$ which describes an aggregate operation, $\mathrm{agg}$ denotes the corresponding operation. Either $\max$, $\min$, $\mathrm{count}$, $\mathrm{sum}$ or $\mathrm{avg}$. \\\hline
$\mathrm{sup}$ & Given $w_{sup}$ describing a superlative, it denotes the corresponding function. Either $\arg\max$ or $\arg\min$. \\\hline
$\mathrm{com}$ & Given $w_{com}$ describing a comparison, it denotes the corresponding relation out of: $<$, $\leq$, $>$, $\geq$, $=$, $\neq$. \\\hline
$\mathrm{ari}$ & Given $w_{ari}$ describing an arithmetic operation, it denotes the corresponding operation out of: $+$, $-$, $*$, $/$. \\\hline
$\mathrm{ground}_{\mathcal{K}}^{\mathrm{e}}(w)$ & Given a natural language phrase $w$, it returns the set of corresponding KB entities, $S_e$. \\\hline
$\mathrm{ground}_{\mathcal{K}}^{\mathrm{r}}(w)$ & Given a natural language phrase $w$, it returns the corresponding KB relation, $r$.  \\\hline
\end{tabular}
\end{center}
\caption{\label{table:grounding_functions} 
Functions used for grounding natural language phrases in numerical operators or KB entities.}
\end{table}

\paragraph{QDMR Operators}\label{subsec:qdmr_operations}
A QDMR step corresponds to one of 13 query operators. We designed the operators to be expressive enough to represent the meaning of questions from a diverse set of datasets (\S\ref{sec:data_collection}).
QDMR assumes an underlying KB, $\mathcal{K}$ which contains all of the entities and relations expressed in its steps. 
A relation, $r$ is a function mapping two arguments to whether $r$ holds in $\mathcal{K}$: $\llbracket r(x,y) \rrbracket_\mathcal{K} \in \{\mathrm{true}, \mathrm{false}\}$. 
The operators operate over: (i) sets of objects $S_o$, where \emph{objects} $o$, are either \emph{numbers} $n$, \emph{boolean} values $b$, or \emph{entities} $e$ in $\mathcal{K}$; (ii) a closed set of phrases $w_{op}$, describing logical operations; (iii) natural language phrases $w$, representing entities and relations in $\mathcal{K}$.
We assume the existence of grounding functions that map a phrase $w$ to concrete constants in $\mathcal{K}$. Table~\ref{table:grounding_functions} describes the aforementioned constructs.
In addition, we define the function $\mathrm{map}_{\mathcal{K}}(S_e, S_o)$ which maps entity $e \in S_e$ to the set of corresponding objects from $S_o$. Each $o \in S_o$ corresponds to an $e \in S_e$ by being contained in the result of a sequence of \texttt{PROJECT} and \texttt{GROUP} operations applied to $e$:\footnote{The sequence of operations $\mathrm{op}_1, \dots, \mathrm{op}_k$ is traced using the references to previous steps in the QDMR structure.}
\begin{multline*}
\mathrm{map}_{\mathcal{K}}(S_e, S_o) =  \{ \langle e,o \rangle \mid e \in S_e, o \in S_o, \\ o \in \mathrm{op}_k \circ...\circ \mathrm{op}_1 (e) \}.
\end{multline*}

We now formally define each QDMR operator and provide concrete examples in Table \ref{table:operators}.

\begin{itemize}[topsep=0pt, itemsep=0pt, leftmargin=.0in, parsep=2pt]
    \item \textbf{\texttt{SELECT:}} Computes the set of entities in $\mathcal{K}$ corresponding to $w$: $\mathbf{select}(w) = \mathrm{ground}_{\mathcal{K}}^{\mathrm{e}}(w)$.
    %\begin{multline*}
    %\mathbf{select}(w) =  \{ e \mid e \in %map_{\mathcal{K}}^{e}(w) \}.
    %\end{multline*}
    \item \textbf{\texttt{FILTER}:} Filters a set of objects so that it follows the condition expressed by $w$:
    \begin{multline*}
    \mathbf{filter}(S_o, w) =  S_o \cap \{ o \mid \llbracket r(e,o) \rrbracket_{\mathcal{K}} \equiv \mathrm{true} \},
    \end{multline*}
    where $r = \mathrm{ground}_{\mathcal{K}}^{\mathrm{r}}(w)$, $e = \mathrm{ground}_{\mathcal{K}}^{\mathrm{e}}(w)  \}$.
    \item \textbf{\texttt{PROJECT:}} Computes the objects that relate to input entities $S_e$ with the relation expressed by $w$,
    \begin{multline*}
    \mathbf{proj}(w, S_e) =  \{ o \mid \llbracket r(e,o) \rrbracket_{\mathcal{K}} \equiv \mathrm{true}, e \in S_e \},
    \end{multline*}
    where $r = \mathrm{ground}_{\mathcal{K}}^{\mathrm{r}}(w)$.
    \item \textbf{\texttt{AGGREGATE}:} The result of applying an aggregate operation: $\mathbf{aggregate}(w_{agg}, S_o) = \{\mathrm{agg}(S_o)\}$.
    \item \textbf{\texttt{GROUP}:} Receives a set of ``keys'', $S_e$ and a set of corresponding ``values'', $S_o$. It outputs a set of numbers, each corresponding to a key $e \in S_e$. Each number results from applying aggregate, $w_{agg}$ to the subset of values corresponding to $e$.
    \begin{multline*}
    \mathbf{group}(w_{agg}, S_o, S_e) = \{ \mathrm{agg}(V_o(e)) \mid e \in S_e \},
    \end{multline*}
    where $V_o(e) = \{o \mid \langle e,o \rangle  \in \mathrm{map}_{\mathcal{K}}(S_e, S_o)\}$.
    \item \textbf{\texttt{SUPERLATIVE}:} Receives entity set $S_e$ and number set $S_n$. Each number $n \in S_n$ is the result of a mapping from an entity $e \in S_e$. It returns a subset of $S_e$ for which the corresponding number is either highest/lowest as indicated by $w_{sup}$.
    \begin{multline*}
    \mathbf{super}(S_e, S_n, w_{sup}) =  \{\mathrm{sup}(\mathrm{map}_{\mathcal{K}}(S_e, S_n)) \}.
    \end{multline*}
    \item \textbf{\texttt{COMPARATIVE}:} Receives entity set $S_e$ and number set $S_n$. Each $n \in S_n$ is the result of a mapping from an $e \in S_e$. It returns a subset of $S_e$ for which the comparison with $n'$, represented by $w_{com}$, holds.
    \begin{multline*}
    \mathbf{comparative}(S_e, S_n, w_{com}, n') =  \{ e \mid \\ \langle e,n \rangle \in \mathrm{map}_{\mathcal{K}}(S_e, S_n), \mathrm{com}(n, n') \equiv \mathrm{true} \}.
    \end{multline*}
    \item \textbf{\texttt{UNION}:} Denotes the union of object sets:
    $\mathbf{union}(S^1_o, S^2_o) = S^1_o \cup S^2_o$.
    \item \textbf{\texttt{DISCARD}:} Denotes the set difference of two objects sets:
    $\mathbf{discard}(S^1_o, S^2_o) =  S^1_o \setminus S^2_o$.
    \item \textbf{\texttt{INTERSECTION}:} Computes the intersection of its entity sets and returns all objects which relate to the entities with the relation expressed by $w$.
    \begin{multline*}
    \mathbf{intersect}(w, S^1_e, S^2_e) = \{ o \mid e \in S^1_e \cap S^2_e, \\ \llbracket r(e,o) \rrbracket_{\mathcal{K}} \equiv \mathrm{true},  r=\mathrm{ground}_{\mathcal{K}}^{\mathrm{r}}(w) \}.
    \end{multline*}
    \item \textbf{\texttt{SORT}:} Orders a set of entities according to a corresponding set of numbers. Each number $n_i$ is the result of a mapping from entity $e_i$.
    \begin{multline*}
    \mathbf{sort}(S^1_e, S^2_n) =  \{ \langle e_{i_1}...e_{i_m}\rangle \mid\\ \langle e_{i_j},n_{i_j} \rangle \in \mathrm{map}_{\mathcal{K}}(S_e, S_n), n_{i_1}\leq...\leq n_{i_m} \}.
    \end{multline*}
    \item \textbf{\texttt{BOOLEAN}:} Returns whether the relation expressed by $w$ holds between the input objects:
    $\mathbf{boolean}(S^1_o, w, S^2_o) = \{ \llbracket r(o_1, o_2) \rrbracket_{\mathcal{K}} \}$, where $r = \mathrm{ground}_{\mathcal{K}}^{\mathrm{r}}(w)$ and $S^1_o$, $S^2_o$ are singleton sets containing $o_1$, $o_2$ respectively.
    \item \textbf{\texttt{ARITHMETIC}:} Computes the application of an arithmetic operation:
    $\mathbf{arith}(w_{ari}, S^1_n, S^2_n) = \{\mathrm{ari}(n_1, n_2)\}$, where $S^1_n$, $S^2_n$ are singleton sets containing $n_1$, $n_2$ respectively.
\end{itemize}

\begin{figure}[t]
    \scriptsize
    \centering
    \begin{tabular}{p{7cm}}
         \hline
         \textbf{Question}: \textit{``The actress that played  Pearl Gallagher on the TV sitcom Different Strokes is also the voice of an animated character that debuted in what year?''}
         \\ \hline
         \textbf{High-level QDMR}: 
         \begin{enumerate}[nosep,after=\strut]
             \item Return the  actress that played Pearl Gallagher on the TV sitcom Different Strokes 
             \item Return the  animated character that {\color{blue} \#1} is the voice of
             \item Return the year that {\color{blue} \#2} debuted in
         \end{enumerate}\vspace{-1em}\vspace{-1em}
         \\ \hline
    \end{tabular}
    \caption{Example of a \textit{high-level} QDMR. Step \#1 merges together \texttt{SELECT} and multiple \texttt{FILTER} steps. }
    \label{figure:high_level_example}
\end{figure}

\paragraph{High-level Decompositions} 
\label{subsec:high_level}
In QDMR, each step corresponds to a single logical operator. In certain contexts, a less granular decomposition might be desirable, where sub-structures containing multiple operators could be collapsed to a single node. This can be easily achieved in QDMR by merging certain adjacent nodes in its DAG structure. When examining existing RC datasets \cite{yang2018HotpotQAAD, dua2019drop}, we observed that long spans in the question often match long spans in the text, due to existing practices of generating questions via crowdsourcing. 
In such cases, decomposing the long spans into multiple steps and having an RC model process each step independently, increases the probability of error. Thus, to promote the usefulness of QDMR for current RC datasets and models, we introduce \emph{high-level} QDMR, by merging the following operators:

\begin{itemize}[topsep=1.5pt, itemsep=3pt, leftmargin=.2in, parsep=2pt]
    \item \textbf{\texttt{SELECT} + \texttt{PROJECT} on named entities:
    %\footnote{We distinguish between \emph{named entities} and entities that should be inferred (e.g., \emph{``the actor portraying Achilles''}).}
    } 
    For the question, \textit{``What is the birthdate of Jane?''} its \textit{high-level} QDMR would be \textit{``return the birthdate of Jane''} as opposed to the more granular, \textit{``return Jane; return birthdate of \#1''}.
    \item \textbf{\texttt{SELECT} + \texttt{FILTER}:} Consider the first step of the example in
    Figure \ref{figure:high_level_example}. It contains both a \texttt{SELECT} operator (\textit{``return actress''}) as well as two \texttt{FILTER} conditions (\textit{``that played...''}, \textit{``on the TV sitcom...''}).
    \item \textbf{\texttt{FILTER} + \texttt{GROUP} + \texttt{COMPARATIVE}:} Certain \textit{high-level} \texttt{FILTER} steps contain implicit grouping and comparison operations. E.g., \textit{``return yard line scores in the fourth quarter; return \#1 that both teams scored from''}. Step \#2 contains an implicit \texttt{GROUP} of team per yard line and a \texttt{COMPARATIVE} returning the lines where exactly two teams scored. 
\end{itemize}

We provide both granular and high-level QDMRs for a random subset of RC questions
%, demonstrating the differences between the granularity of decompositions 
(see Table \ref{table:datasets}).
%In contexts other than RC, other levels of granularity might prove to be beneficial.
The concrete utility of \textit{high-level} QDMR to open-domain QA is presented in \S\ref{sec:application_qa}.

\section{Data Collection}
\label{sec:data_collection}

Our annotation pipeline for generating \datasetname{} consisted of three phases. %First, we collected a large number of complex questions from existing benchmarks over multiple modalities. 
First, we collected complex questions from existing QA benchmarks.
Second, we crowdsourced the QDMR annotation of these questions. Finally, we validated worker annotations in order to maintain their quality.

\paragraph{Question Collection}
Questions in \datasetname{} were randomly sampled from ten QA datasets over the following tasks (Table~\ref{table:datasets}):
\begin{itemize}[topsep=0pt, itemsep=0pt, leftmargin=.2in, parsep=0pt]
    \item \textbf{Semantic Parsing:}
    Mapping natural language utterances into formal queries, to be executed on a target KB  \cite{price1990atis,zelle96geoquery,Li2014NaLIRAI,Yu2018SpiderAL}.
    \item \textbf{Reading Comprehension (RC):}
    Questions that require understanding of a text passage by reasoning over multiple sentences \cite{talmor2018web, yang2018HotpotQAAD,dua2019drop, Abujabal2018ComQAAC}.
    \item \textbf{Visual Question Answering (VQA):}
    Questions over images that require both visual and numerical reasoning skills \cite{johnson2017clevr, suhr2018corpus}.
\end{itemize}

All questions collected were composed by human annotators.\footnote{Except for \textsc{ComplexWebQuestions} (\textsc{CWQ}), where annotators paraphrased automatically generated questions.} \textsc{HotpotQA} questions were all sampled from the \textit{hard} split of the dataset.

\begin{table}[t!]
\begin{center}
\tiny
\begin{tabular}{p{1cm}|p{3cm}|p{0.8cm}|p{0.8cm}}
\hline \bf Dataset & \bf Example & \bf Original & \bf \datasetname{} \\ \hline
\sc Academic (DB) & \it Return me the total citations of all the papers in the VLDB conference. & 195 & 195 \\ \hline
\sc ATIS (DB) & \it What is the first flight from Atlanta to Baltimore that serves lunch? & 5,283 & 4,906 \\ \hline
\sc GeoQuery (DB) & \it How high is the highest point in the largest state? & 880 & 877 \\ \hline
\sc Spider (DB) & \it How many transactions correspond to each invoice number? & 10,181 & 7,982 \\ \hline
\sc CLEVR-humans (Images) & \it What is the number of cylinders divided by the number of cubes? & 32,164 & 13,935 \\ \hline
\sc NLVR2 (Images) & \it If there are only two dogs pulling one of the sleds? & 29,680 & 13,517 \\ \hline
\sc ComQA (Text) & \it What was Gandhi's occupation before becoming a freedom fighter? & 11,214 & 5,520 \\ \hline
\sc CWQ (Text) & \it Robert E Jordan is part of the organization started by whom? & 34,689 & 2,988, 2,991$_{\textit{high}}$ \\ \hline
\sc DROP (Text) & \it Approximately how many years did the churches built in 1909 survive? & 96,567 & 10,230, 10,262$_{\textit{high}}$ \\ \hline
\sc HotpotQA-hard (Text) &  \it Benjamin Halfpenny was a footballer for a club that plays its home matches where? & 23,066 & 10,575$_{\textit{high}}$ \\ \hline
\multicolumn{3}{l|}{\bf Total:}   & \bf 83,978 \\ \hline
\end{tabular}
\end{center}
\caption{\label{table:datasets} The QA datasets in \datasetname{}. Lists the number 
  of examples in the original dataset and in \datasetname{}. Numbers of \textit{high-level} QDMRs are denoted by \textit{high}.
}
\end{table}

\begin{figure}[t]
\centering
  \includegraphics[trim={0cm 4cm 14cm 0cm}, clip, width=7.5cm,height=5.8cm]{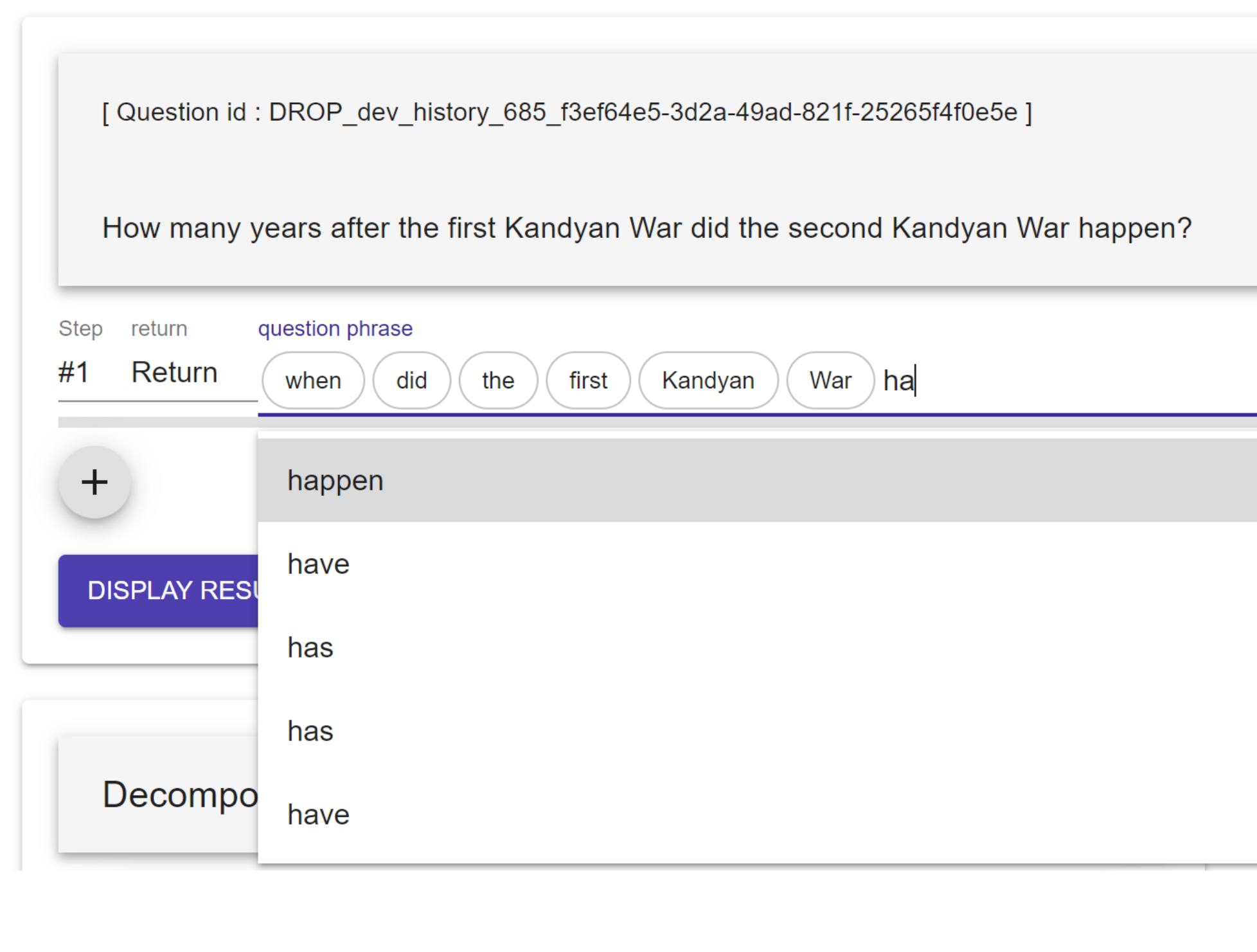}
\caption{User interface for decomposing a complex question that uses a closed lexicon of tokens.}
\label{fig:interface}
\end{figure}

\paragraph{QDMR Annotation}
A key question is whether it is possible to train non-expert annotators to produce high-quality QDMRs. We designed an annotation interface (Figure~\ref{fig:interface}), where workers are first given explanations and examples on how to identify and phrase each of the operators in Table~\ref{table:operators}. Then, workers decompose questions into a list of steps, where they are only allowed to use words from a lexicon $L_x$, which contains: (a) words appearing in the question (or their automatically computed inflections), (b) words from a small pre-defined list of 66 function word such as, \emph{`if'}, \emph{`on'}, \emph{`for each'}, or (c) \emph{reference tokens} that refer to the results of a previous step. This ensures that the language used by workers is consistent across examples, while being expressive enough for the decomposition. Our annotation interface presents workers with the question only, so they are agnostic to the original modality of the question. The efficacy of this process is explored in \S\ref{subsec:quality_analysis}. 

We used Amazon Mechanical Turk to crowdsource QDMR annotation. In each task, workers decomposed a single question, paying them \$0.4, which amounts to an average pay of \$12 per hour.
Overall, we collected 83,978 examples using 64 distinct workers. The dataset was partitioned into train/development/test sets following the partitions in the original datasets. 
During partition, we made sure that development and test samples do not share the same context.

\paragraph{Worker Validation}
To ensure worker quality, we initially published qualification tasks, open to all United States' workers. The task required workers to carefully review the annotation instructions and decompose 10 example questions. The examples were selected so that each QDMR operation should appear in at least one of their decompositions (Table \ref{table:operators}). In total, 64 workers were able to correctly decompose at least 8 examples and were qualified as annotators.
To validate worker performance over time, we conducted random validations of annotations. Over 9K annotations were reviewed by experts throughout the annotation process. Only workers that consistently produced correct QDMRs for at least 90\% of their tasks were allowed to continue as annotators.

\section{Dataset Analysis}
\label{sec:data_analysis}
This section examines the properties of collected QDMRs in \datasetname{} and analyzes their quality.

%First, we describe the data composition with regards to the different modalities and datasets. We then analyze the prevalence of the various QDMR operators in the dataset. We conclude by providing a post-hoc estimation of the quality of QDMR annotations and validating their consistency.

\subsection{Quantitative Analysis}
\label{subsec:data_statistics}
Overall, \textsc{Break} contains 83,978 decompositions, including 60,150 QDMRs and 23,828 examples with \textit{high-level} QDMRs, which are exclusive to text modalities.
Table \ref{table:datasets} shows data is proportionately distributed between questions over structured (DB) and unstructured modalities (text, images). 
%This diversity sets \datasetname{} as a general purpose benchmark for question decomposition.

%\subsection{QDMR Structure Analysis}
\label{subsec:qdmr_structure_analysis} 
%The following section provides further analysis into the properties of QDMRs in \textsc{Break}. 
The distribution of QDMR operators is presented in Table \ref{table:operator_prevalence}, detailing the prevalence of each query operator\footnote{Regarding the three merged operators of \emph{high-level} QDMRs (\S\ref{subsec:high_level}), the first two operators are treated as \texttt{SELECT}, while the third is considered a \texttt{FILTER}.} (we automatically compute this distribution, as explained in \S\ref{subsec:annotation_consistency}). \texttt{SELECT} and \texttt{PROJECT} are the most common operators. Additionally, at least 10\% of QDMRs contain operators such as \texttt{GROUP} and \texttt{COMPARATIVE} which entail complex  reasoning, in contrast to \emph{high-level} QDMRs, where such operations are rare.
This distinction sheds light on the reasoning types required for answering RC datasets (\emph{high-level} QDMR) compared to more structured tasks (QDMR).

Table \ref{table:qdmr_lengths} details the distribution of QDMR sequence length. Most decompositions in QDMR include 3-6 steps, while \emph{high-level} QDMRs are much shorter, as a single \texttt{SELECT} often finds an entity described by a long noun phrase (see \S\ref{subsec:high_level}).

\subsection{Quality Analysis}\label{subsec:quality_analysis}
We describe the process of estimating the \emph{correctness} of collected QDMR annotations.
%\jb{why not have something like: Since there are multiple correct QDMR annotations for every question, measuring inter-annotator agreement is problematic. Thus, similar to past work, ....}
Similar to previous works \cite{Yu2018SpiderAL, kwiatkowski2019natural} we use expert judgements, where the experts had prepared the guidelines for the annotation task. Given a question and its annotated QDMR, $(q, \decomplist)$ the expert determines the correctness of $\decomplist$ using one of the following categories:
\begin{itemize}[topsep=0pt, itemsep=0pt, leftmargin=.2in, parsep=0pt]
    \item Correct ($\mathcal{C}$): If $\decomplist$ constitutes a list of QDMR operations that lead to correctly answering $q$.
    \item Granular ($\mathcal{C_{G}}$): If $\decomplist$ is correct and none of its operators can be further decomposed.\footnote{For \emph{high-level} QDMRs, the merged operators (\S\ref{subsec:high_level}) are considered to be fully decomposed.}
    \item Incorrect ($\mathcal{I}$): If $\decomplist$ is in neither $\mathcal{C}$ nor $\mathcal{C_{G}}$.
\end{itemize}

Examples of these expert judgements are shown in Figure \ref{figure:expert_evaluation}. To estimate expert judgement of correctness, we manually reviewed a random sample of 500 QDMRs from \datasetname{}. We classified 93.8\% of the samples in $\mathcal{C_{G}}$ and another 3.6\% in $\mathcal{C}$. Thus, 97.4\% of the samples constitute a correct decomposition of the original question. Workers have somewhat struggled with decomposing superlatives (e.g., \textit{``biggest sphere''}), as evident from the first question in Figure \ref{figure:expert_evaluation}. Collected QDMRs displayed similar estimates of $\mathcal{C}$, $\mathcal{C_{G}}$ and $\mathcal{I}$, regardless of their modality (DB, text or image).

\begin{figure}[t!]
    \centering
    \scriptsize
    \begin{tabular}{p{7cm}}
    \hline
        \textbf{Question 1:} \textit{``What color is the biggest sphere in the picture?''}\\
        \textbf{QDMR:} (1) Return spheres; (2) Return {\color{blue} \#1} that is the biggest; (3) Return color of {\color{blue} \#2}. \\
        \textbf{Expert Judgement:} $\mathcal{C}$. Correct, but not fully decomposed. Step \#2 should be broken down to: (2) Return size of {\color{blue} \#1}; (3) Return {\color{blue} \#1} where {\color{blue} \#3} is highest.\\
        \hline 
        \textbf{Question 2:} \textit{``What is the full name of each car maker, along with its id and how many models it produces?''}\\
        \textbf{QDMR:} (1) Return car makers; (2) Return models of {\color{blue} \#1}; (3) Return number of {\color{blue} \#2} for each {\color{blue} \#1}; (4) Return full names of {\color{blue} \#1}; (5) Return ids of {\color{blue} \#1}; (6) Return {\color{blue} \#4} , {\color{blue} \#5} , {\color{blue} \#3}.\\
        \textbf{Expert Judgement:} $\mathcal{C_{G}}$. Correct and fully decomposed.\\
        \hline 
        \textbf{Question 3:} \textit{``Show the names and locations of institutions that are founded after 1990 and have the type Private.''}\\
        \textbf{QDMR:} (1) Return institutions; (2) Return {\color{blue} \#1} founded after 1990; (3) Return types of {\color{blue} \#1}; (4) Return {\color{blue} \#1} where {\color{blue} \#3} is Private; (5) Return {\color{blue} \#2} , {\color{blue} \#4}; (6) Return names of {\color{blue} \#5}; (7) Return locations of  {\color{blue} \#5}; (8) return {\color{blue} \#6} ,  {\color{blue} \#7}. \\
        \textbf{Expert Judgement:} $\mathcal{I}$. Incorrect, as step \#5 returns institutions that were either founded after 1990, or are Private.\\
        \hline 
    \end{tabular}
    \caption{Examples and justifications of expert judgement on collected QDMRs in \datasetname{}.}
    \label{figure:expert_evaluation}
\end{figure}

\subsection{Annotation Consistency}\label{subsec:annotation_consistency}

As QDMR is expressed using natural language, it introduces variability into its annotations. We wish to validate the \emph{consistency} of collected QDMRs, i.e., whether we can correctly infer the formal QDMR operator ($f^i$) and its arguments from each step ($s^i$). To infer these formal representations, we developed an algorithm that goes over the QDMR structure step-by-step, and for each step $s^i$, uses a set of predefined templates to identify $f^i$ and its arguments, expressed in $s^i$. This results in an execution graph (Figure \ref{fig_decomposition_dag}), where the execution result of a parent node serves as input to its child.
Figure \ref{figure:motivation} presents three QDMR decompositions along with the formal graphs output by our algorithm (lower box). Each node lists its operator (e.g., \texttt{GROUP}), its \textit{constant input} listed in brackets (e.g., \texttt{count}) and its \textit{dynamic input} which are the execution results of its parent nodes. 

Overall, 99.5\% of QDMRs had all their steps mapped into pseudo-logical forms by our algorithm. 
%We then wished to assert that the logical forms indeed had their steps mapped into the correct QDMR operator. 
To evaluate the correctness of the mapping algorithm, we randomly sampled 350 logical forms, and examined the structure of the formulas, assuming that words copied from the question correspond to entities and relations in an idealized KB (see \S\ref{sec:decomposition_formalism}). Of this sample, 99.4\% of its examples had all of their steps, $s^i$, correctly mapped to the corresponding $f^i$. Overall, 93.1\% of the examples were of fully accurate logical forms, with errors being due to QDMRs that were either incorrect or not fully decomposed ($\mathcal{I}$, $\mathcal{C}$ in \S\ref{subsec:quality_analysis}). Thus, a rule-based algorithm can map more than 93\% of the annotations into a correct formal representation.
This shows our annotators produced consistent and high-quality QDMRs. Moreover, it suggests that non-experts can annotate questions with pseudo-logical forms, which can be used as a cheap intermediate representation for semantic parsers \cite{yih2016value}, further discussed in \S\ref{subsec:application_low}.

\begin{table}[t]
\begin{center}
\scriptsize
\begin{tabular}{l | c c}
\hline \bf Operator & \bf QDMR & \bf QDMR$_{\textit{high}}$ \\ \hline
\bf\texttt{SELECT} & 100\% &  100\% \\ 
\bf\texttt{PROJECT} & 69.0\% &  35.6\% \\ 
\bf\texttt{FILTER} & 53.2\% &  15.3\% \\ 
\bf\texttt{AGGREGATE} & 38.1\% &  22.3\% \\ 
\bf\texttt{BOOLEAN} & 30.0\% &  4.6\% \\ 
\bf\texttt{COMPARATIVE} & 17.0\% &  1.0\% \\ 
\bf\texttt{GROUP} & 9.7\% &  0.7\% \\ 
\bf\texttt{SUPERLATIVE} & 6.3\% &  13.0\% \\ 
\bf\texttt{UNION} & 5.5\% &  0.5\% \\ 
\bf\texttt{ARITHMETIC} & 5.4\% &  11.2\% \\ 
\bf\texttt{DISCARD} & 3.2\% &  1.2\% \\ 
\bf\texttt{INTERSECTION} & 2.7\% &  2.8\% \\ 
\bf\texttt{SORT} & 0.9\% &  0.0\% \\ \hline
Total & 60,150 & 23,828 \\ \hline
\end{tabular}
\end{center}
\caption{\label{table:operator_prevalence} 
Operator prevalence in \datasetname{}. Lists the percentage of QDMRs where the operator appears.
}
\end{table}

\begin{table}[t]
\begin{center}
\scriptsize
\begin{tabular}{r | c c}
\hline \bf Steps & \bf QDMR & \bf QDMR$_{\textit{high}}$ \\ \hline
1-2 & 10.7\% & 59.8\%  \\ 
3-4 & 44.9\% &  31.6\%  \\ 
5-6 & 27.0\% &  7.9\%  \\ 
7-8 & 10.1\% &  0.6\%  \\ 
9+ & 7.4\% &  0.2\%  \\ \hline

\end{tabular}
\end{center}
\caption{\label{table:qdmr_lengths} 
The distribution over QDMR sequence length.
}
\end{table}

\section{QDMR for Open-domain QA}
\label{sec:application_qa}
%We present an application of QDMR to open-domain QA. In addition, we demonstrate the potential of QDMR as an intermediate representation for semantic parsing.

%The structure of QDMR facilitates its mapping into formal queries.  In addition, we use \datasetname{} to bootstrap a QA model trained on single-hop questions to handle questions involving multi-hop reasoning.

A natural setup for QDMR is in answering complex questions that require multiple reasoning steps. 
We compare models that exploit question decompositions to baselines that do not.
We use the open-domain QA (``full-wiki")  setting of the \textsc{HotpotQA} dataset \cite{yang2018HotpotQAAD}: Given a question, the QA model retrieves the relevant Wikipedia paragraphs and answers the question using these paragraphs.
%%% TW - for ComplexWebQuestions Experiments
%We use the open-domain QA ``full-wiki"  settings of the \textsc{HotpotQA} and \textsc{CWQ} datasets \cite{yang2018HotpotQAAD, talmor2018web}: Given a question, the QA model must retrieve relevant Wikipedia paragraphs and answer the question using the paragraphs.

\subsection{Experimental setup}\label{subsec:application_high}
%A natural setup for QDMR is in answering complex questions that require multiple reasoning steps. 
%We compare a model that exploits question decompositions to baselines that do not. 
%The utility of QDMR is examined in the open-domain QA settings of the \textsc{HotpotQA} and \textsc{CWQ} datasets \cite{yang2018HotpotQAAD, talmor2018web}. We use the ``full-wiki'' setting: Given a question, the ``full-wiki' QA model must learn to retrieve relevant paragraphs from Wikipedia and then answer the question using the retrieved context.

We compare \textsc{BreakRC}, a model that utilizes question decomposition to \textsc{BERTQA}, a standard QA model, based on BERT \cite{devlin2018bert}, and present \textsc{Combined}, an approach that enjoys the benefits of both models.

\paragraph{\textsc{BreakRC}}
Algorithm \ref{alg:break_rc} describes the \textsc{BreakRC} model which uses high-level QDMR structures for answering open-domain multi-hop questions. We assume access to an IR model and an RC model, and denote by $\textsc{Answer}(\cdot)$ a function that takes a question as input, runs the IR model to obtain paragraphs, and then feeds those paragraphs as context for an RC model that returns a distribution over answers.

Given an input  QDMR, $\decomplist = \langle s^1, ..., s^n \rangle$, iterate over $\decomplist$ step-by-step and perform the following. First, we extract the operation (line~\ref{line:op_ext}) and the previous steps referenced by $s^i$ (line~\ref{line:ref_ext})
. Then, we compute the answer to $s^i$ conditioned on the extracted operator. For \texttt{SELECT} steps, we simply run the $\textsc{Answer}(\cdot)$ function. For \texttt{PROJECT} steps, we substitute the reference to the previous step in $s^i$ with its already computed answer, and then run $\textsc{Answer}(\cdot)$. For \texttt{FILTER} steps,\footnote{\texttt{INTERSECTION} steps are handled in a manner similar to \texttt{FILTER}, but we omit the exact description for brevity.} we use a simple rule to extract a ``normalized question'',  $\hat{s}^i$ from $s^i$ and get an intermediate answer $ans_\text{tmp}$ with $\textsc{Answer}(\hat{s}^i)$. We then ``intersect'' $ans_\text{tmp}$ with the referenced answer by multiplying the probabilities provided by the RC model and normalizing. For \texttt{COMPARISON} steps, we compare, with a discrete operation, the numbers returned by the referenced steps. The final answer is the highest probability answer of step $s^n$.

As our IR model we use bigram TF-IDF, proposed by \newcite{chen2017reading}. Since the RC model is run on single-hop questions, we use the BERT-based RC model from \newcite{min2019multi}, trained solely on SQuAD \cite{rajpurkar2016squad}.

\begin{algorithm}[t]
\scriptsize
\caption{\textsc{BreakRC}}
\begin{algorithmic}[1]
\Procedure{BreakRC}{$\decomplist$ : QDMR}
    \State $ansrs\gets []$
    \For{$s^i$ in $\decomplist = \langle s^1, ..., s^n \rangle$}
        \State $op\gets\textsc{OpType}(s^i)$ \label{line:op_ext}
        \State $refs \gets\textsc{ReferencedSteps}(s^i)$ \label{line:ref_ext}
        \If{$op$ is \texttt{SELECT}}
          \State $ans\gets\textsc{Answer}(s^i)$
        \ElsIf{$op$ is \texttt{FILTER}}
          \State $\hat{s}^i\gets\textsc{ExtractQuestion}(s^i)$
          \State $ans_\text{tmp}\gets\textsc{Answer}(\hat{s}^i)$
          \State $ans\gets\textsc{Intersect}(ans_\text{tmp}, ansrs[refs[0]])$
        \ElsIf{$op$ is \texttt{COMPARISON}}
          \State $ans\gets\textsc{CompareSteps}(refs, $\decomplist$)$
        \Else \Comment{$op$ is \texttt{PROJECT}}
          \State $\hat{s}^i\gets\textsc{SubstituteRef}(s^i, ansrs[refs[0]])$
          \State $ans\gets\textsc{Answer}(\hat{s}^i)$
        \EndIf
        \State $ansrs[i]\gets ans$
    \EndFor
    \State \textbf{return} $ansrs[n]$ 
    %\State \textbf{return} $\langle answer, contexts \rangle$ 
\EndProcedure
\end{algorithmic}
\label{alg:break_rc}
\end{algorithm}

\paragraph{\textsc{BERTQA} Baseline}
As \textsc{BreakRC} exploits question decompositions, we compare it with a model that does not. \textsc{BERTQA} receives as input the original natural language question, $x$. It uses the same IR model as \textsc{BreakRC} to retrieve paragraphs for $x$. For a fair comparison, we set its number of retrieved paragraphs such that it is identical to \textsc{BreakRC} (namely, 10 paragraphs for each QDMR step that involves IR).
Similar to \textsc{BreakRC}, retrieved paragraphs are fed to a pretrained BERT-based RC model \cite{min2019multi} to answer $x$. 
In contrast to \textsc{BreakRC}, that is trained on \textsc{SQuAD}, \textsc{BERTQA} is trained on the target dataset (\textsc{HotpotQA}), giving it an advantage over \textsc{BreakRC}.

\begin{figure}[t!]
    \centering
    \scriptsize
    \begin{tabular}{p{7cm}}
    \hline
        \textbf{Project Question:} \textit{``The actor of Professor Sprout in Harry Potter acted as a mother in a sitcom. In that sitcom, what was the daughter's name?''}\\
        \textbf{Gold Decomposition:} \\
        1. Return actor of Professor Sprout In Harry Potter \\
        2. Return sitcom that {\color{blue} \#1} acted as a mother in \\
        3. Return daughter name in {\color{blue} \#2} \\ \hline 
        \textbf{Comparison Question:} \textit{``Which 1970's film was released first, Charley and the Angel or The Boatniks?''}\\
        \textbf{Gold Decomposition:}  \\
        1. Return when Charley and the Angel was released \\
        2. Return when The Boatniks was released \\
        3. Return which is the lowest of  {\color{blue} \#1} , {\color{blue} \#2} \\ \hline 
    \end{tabular}
    \caption{Examples of \texttt{PROJECT} and \texttt{COMPARISON} questions in \textsc{HotpotQA} (\emph{high-level} QDMR).}
    \label{figure:hotpotqa_decompositions}
\end{figure}

\paragraph{A \textsc{Combined} Approach}\label{subsubsec:application_high_combined}
Last, we present an approach that combines the strengths of \textsc{BreakRC} and \textsc{BERTQA}.
In this approach, we use the QDMR decomposition to improve \emph{retrieval} only.
Given a question $x$ and its QDMR $\decomplist$, we run \textsc{BreakRC} on $\decomplist$, but in addition to storing \emph{answers}, we also store all the \emph{paragraphs} retrieved by the IR model. We then run \textsc{BERTQA} on the question $x$ and the top-10 paragraphs retrieved by \textsc{BreakRC}, sorted by their IR ranking. This approach resembles that of \citet{qi-etal-2019-answering}.

The advantage of \textsc{Combined} is that we do not need to develop an answering procedure for each QDMR operator separately, which involves different discrete operations such as comparison and intersection. Instead, we use \textsc{BreakRC} to retrieve contexts, and an end-to-end approach to learn how to answer the question directly. This can often handle operators not implemented in \textsc{BreakRC}, like \texttt{BOOLEAN} and \texttt{UNION}.

%The \textsc{Combined} approach, receives as input the original question, $x$ and its QDMR $\decomplist$. It first runs \textsc{BreakRC} on $\decomplist$. However, instead of returning the answer of \textsc{BreakRC}, it only uses its retreived contexts. Taking as context the top 10 paragraphs returned by the union of all previous contexts. Last, \textsc{Combined} runs the \textsc{BERTQA} RC model on $x$ and the \textsc{BreakRC} context to return the answer.

%By utilizing \textsc{BERTQA}, \textsc{Combined} alleviates the need of \textsc{BreakRC} to manually handle all the discrete QDMR operation types. E.g., our \textsc{BreakRC} implementation does not handle \texttt{BOOLEAN}, \texttt{UNION} and \texttt{DISCARD} steps, which account for over 10\% of the examples. In addition, \textsc{Combined} leverages the structure of QDMRs to correctly retrieve the context for bridge questions. wha

\paragraph{\textsc{Dataset}} To evaluate our models,
we use all 2,765 QDMR annotated examples of the \textsc{HotpotQA} development set found in \datasetname{}. \texttt{PROJECT} and \texttt{COMPARISON} type questions account for 48\% and 7\% of examples respectively.
%%% TW - For ComplexWebQuestions experiments:
%For CWQ, we use all 475 annotated examples of its development set. 
%In CWQ, we restrict the IR model to Wikipedia, unlike the original setting in \citet{talmor2018web} that utilizes web search results.

\subsection{Results}\label{subsubsec:application_high_experiments}

Table \ref{table:hotpotqa_results} shows model performance on \textsc{HotpotQA}. We report EM and F$_1$ using the official \textsc{HotpotQA} evaluation script. IR, measures the percentage of examples in which the IR model successfully retrieved \emph{both} of the ``gold paragraphs'' necessary for answering the multi-hop question.
%We present the models performance over \textsc{HotpotQA} and \textsc{CWQ} in Table \ref{table:hotpotqa_results}. 
To assess the potential utility of QDMR, we report results for $\textsc{BreakRC}^{\textsc{G}}$, which uses \emph{gold} QDMRs, and $\textsc{BreakRC}^{\textsc{P}}$, which uses QDMRs predicted by a \copynet{} parser (\S\ref{subsec:baselines}). 

%As \textsc{BreakRC} depends on question decompositions as input, we experiment with three separate decomposition types. First, the $\textsc{BreakRC}^{\textsc{G}}$ model uses the ``gold decompositions'' found in \datasetname{}. Additionally, we provide results on predicted QDMR structures ($\textsc{BreakRC}^{\textsc{P}}$), using the \copynet{} parser described in \S\ref{subsec:baselines}, trained on high-level QDMRs.
%We view $\textsc{BreakRC}^{\textsc{G}}$ as a reasonable upper bound 
%Results using ``gold'' QDMRs serve as a plausible upper bound on performance, assuming a quality QDMR parser (\S\ref{sec:question_decomposition}).
%Results using ``gold'' QDMRs serve as a plausible upper bound on performance, assuming a quality QDMR parser (\S\ref{sec:question_decomposition}).

Retrieving paragraphs with decomposed questions substantially improves the IR metric from 46.3 to 59.2 ($\textsc{BreakRC}^{\textsc{G}}$), or 52.5 ($\textsc{BreakRC}^{\textsc{P}}$). This leads to substantial gains in EM and F$_1$ for $\textsc{Combined}^{\textsc{G}}$ (43.3 to 52.4) and $\textsc{Combined}^{\textsc{P}}$ (43.3 to 49.3). The EM and F$_1$ of $\textsc{BreakRC}^{\textsc{G}}$ are only slightly higher than \textsc{BERTQA} since \textsc{BreakRC} does not handle certain operators, such as \texttt{BOOLEAN} steps (9.4\% of the examples).

% \rev{
% Unsurprisingly, the performance of \textsc{Combined} on the annotated dev sets is significantly higher than both approaches. By leveraging the structure of QDMR steps, \textsc{Combined} is able utilize \textsc{BreakRC} to retrieve the correct IR context. Having received improved context (the ``IR'' performance), allows \textsc{BERTQA} to directly retrieve the correct answer. This concretely demonstrates the quality of decompositions in \datasetname{}, enabling to significantly boost performance (XX to XX F1). The relatively low performance of \textsc{BreakRC} stems from unhanding discrete QDMR steps other than those in Algorithm \ref{alg:break_rc}. E.g., on questions containing \texttt{BOOLEAN} steps (9.4\% of examples) it achieves zero performance, as its RC model was not trained on boolean questions. 
% }

The majority of questions in \textsc{HotpotQA} combine \texttt{SELECT} operations with either \texttt{PROJECT} (also called ``bridge'' questions), \texttt{COMPARISON}, or \texttt{FILTER}. \texttt{PROJECT} and \texttt{COMPARISON} questions (Figure \ref{figure:hotpotqa_decompositions}) were shown to be less susceptible to reasoning shortcuts, i.e. they necessitate multi-step reasoning \cite{Chen2019UnderstandingDD, jiang-bansal-2019-avoiding, min2019compositional}. In Table \ref{table:hotpotqa_results_operators} we report \textsc{BreakRC} results on these question types, where it notably outperforms \textsc{BERTQA}.

%\textsc{BreakRC} is especially suitable for questions that involve \texttt{SELECT} and \texttt{PROJECT} operations (termed \emph{Bridge} questions in prior work), since IR is difficult

%In Table \ref{table:hotpotqa_results_operators} we further break down model performance based different question types found in \textsc{HotpotQA}. We focus on a particular subset of ``bridge'' and ``comparison'' questions which require complex reasoning capabilities.\footnote{For this experiment we consider compositional bridge questions of the type \emph{``What is the nationality of the director of Titanic?''}, since those have been shown to be less susceptible to ``reasoning shortcuts'' \cite{Chen2019UnderstandingDD, jiang-bansal-2019-avoiding, min2019compositional}.} Example bridge and comparison questions are shown in Figure \ref{figure:hotpotqa_decompositions}. On these question types \textsc{BreakRC} significantly outperforms the \textsc{BERTQA}, further indicating the utility of \datasetname{} decompositions.

\paragraph{Ablations}\label{subsubsec:application_high_ir_baselines}
%\rev{We now examine whether using QDMR was indeed the cause for the \textsc{BreakRC} improvements, as seen in Tables  \ref{table:hotpotqa_results}, \ref{table:hotpotqa_results_operators}.}

%paragraph{IR} 
In \textsc{BreakRC}, multiple IR queries are issued, one at each step. To examine whether these multiple queries were the cause for performance gains, we built \textsc{IR-NP}: A model that issues multiple IR queries, one for each noun phrase in the question. Similar to \textsc{Combined}, the question and union of retrieved paragraphs are given as input to \textsc{BERTQA}. We observe that \textsc{Combined} substantially outperforms \textsc{IR-NP}, indicating that the structure of QDMR, rather than multiple IR queries, has led to improved performance.\footnote{Issuing an IR query over each ``content word'' in the question, instead of each noun phrase, led to poor results.}

%The IR component in \textsc{BreakRC} issues a separate retrieval query for each QDMR step. A reasonable question is whether these multiple TF-IDF queries were the primary cause for the \textsc{BreakRC} gains, rather than its use of QDMR. We put this claim to the test by building an IR baseline that mimics QDMR by issuing multiple TF-IDF queries and taking the union over their results. The \textsc{IR-NP} baseline performs constituent parsing over the original question, then queries each noun phrase constituent separately. Similar to \textsc{Combined}, the retrieved context is then issued to \textsc{BERTQA} along with the original question. Results in Table \ref{table:hotpotqa_results} show that \textsc{IR-NP} performance falls short of \textsc{BreakRC}. This further indicates that the structure of QDMR steps, rather than multiple TF-IDF queries, has led to the \textsc{BreakRC} scores.

To test whether QDMR is better than a simple rule-based decomposition algorithm, we developed a model that decomposes a question by applying a set of predefined rules over the dependency tree of the question (full details in \S\ref{subsec:baselines}). \textsc{Combined} and \textsc{BreakRC} were compared to $\textsc{Combined}^{\textsc{R}}$ and $\textsc{BreakRC}^{\textsc{R}}$ which use the rule-based decompositions. We observe that QDMR lead to substantially higher performance when compared to the rule-based decompositions.

\begin{table}[t!]
    \centering
    \scriptsize
    \begin{tabular}{l | c c c}
    \hline 
        \multirow{2}{*}{Model} & \multicolumn{3}{c}{\textsc{HotpotQA}}\\

         & EM & F$_{1}$ & IR  \\\hline \hline
         \textsc{BERTQA} & 33.6 & 43.3 & 46.3  \\
         $\textsc{BreakRC}^{\textsc{P}}$ & 28.8 & 37.7 & 52.5  \\
         $\textsc{BreakRC}^{\textsc{G}}$ & 34.6 & 44.6 & \bf 59.2  \\
         $\textsc{Combined}^{\textsc{P}}$ & 38.3 & 49.3 & 52.5 \\
         $\textsc{Combined}^{\textsc{G}}$ & \bf 41.2 & \bf 52.4 & \bf 59.2  \\\hline
         $\textsc{IR-NP}$  & 31.7 & 41.2 & 40.8  \\
         $\textsc{BreakRC}^{\textsc{R}}$ & 18.9 & 26.5 & 40.3  \\
         $\textsc{Combined}^{\textsc{R}}$ & 32.7 & 42.6 & 40.3 \\
    \hline
    \end{tabular}
    \caption{Open-domain QA results on \textsc{HotpotQA}.}
    \label{table:hotpotqa_results}
\end{table}

\begin{table}[t!]
    \centering
    \scriptsize
    \begin{tabular}{l | c c c | c c c}
    \hline 
        \multirow{2}{*}{Model} & \multicolumn{3}{c}{\textsc{Project}} & \multicolumn{3}{c}{\textsc{Comparison}} \\

         & EM & F$_{1}$ & IR & EM & F$_{1}$ & IR  \\\hline \hline
         \textsc{BERTQA} & 22.8 & 31.0 & 31.6 & 42.9 & 51.7 &  75.8 \\
         $\textsc{BreakRC}^{\textsc{P}}$ & 25.4 & 33.7 & 52.9 & 34.7 & 50.4 & 68.9 \\
         $\textsc{BreakRC}^{\textsc{G}}$ & \bf 32.2 & \bf 41.9 & \bf 59.8 & \bf 44.5 & \bf 57.6 & \bf 78.0 \\
    \hline
    \end{tabular}
    \caption{Results on \texttt{PROJECT} and \texttt{COMPARISON} questions from \textsc{HotpotQA} development set.}
    \label{table:hotpotqa_results_operators}
\end{table}

\section{QDMR for Semantic Parsing}
\label{subsec:application_low}
%In this section we discuss the potential of QDMR for semantic parsing, and the relation between QDMR and executable semantic parsing. 
%We discuss the potential use of QDMR in semantic parsing tasks and present its limitations compared to fully executable queries.  

As QDMR structures can be easily annotated at scale, a natural question is how far are they from fully executable queries (known to be expensive to annotate). As shown in \S\ref{subsec:annotation_consistency}, QDMRs can be mapped to \emph{pseudo-logical forms} with high precision (93.1\%) by extracting formal operators and arguments from their steps. The pseudo-logical form differs from an executable query in the lack of grounding of its arguments (entities and relations) in KB constants. This stems from the design of QDMR as a domain-agnostic meaning representation (\S\ref{sec:decomposition_formalism}). QDMR abstracts away from a concrete KB schema by assuming an underlying ``idealized'' KB, which contains all of its arguments.

Thus, QDMR can be viewed as an \emph{intermediate representation} between a natural language question and an executable query. 
%It serves as a decomposition of the original question while lacking the grounding in the KB constants. 
Such intermediate representations have already been discussed in prior work on semantic parsing. \citet{kwiatkowski2013scaling} and \citet{choi-etal-2015-scalable} 
used \emph{underspecified logical forms} as an intermediate representation.
%have utilized a two stage approach to learn domain independent semantic parsers using \emph{underspecified logical forms}. Their work was further expanded to handle incomplete KBs  \cite{choi-etal-2015-scalable}. 
\citet{GuoIRNet2019} proposed a two-stage approach, separating between learning an intermediate text-to-SQL representation and the actual mapping to schema items. 
%Neural Modular Networks \cite{andreas2016modular, hu2017learning, jiang2019self} have also been known to separate the semantic question representation, or ``layout'' from the actual question execution through context dependent ``modules''. 
Works in the database community have particularly targeted the mapping of intermediate query representations into DB grounded queries, using schema mapping and join path inference \cite{androutsopoulos95nlidb, li2014schema,baik2019bridging}. We argue that QDMR
can be used as an easy-to-annotate representation in such semantic parsers, bridging between natural language and full logical forms.

\section{QDMR Parsing}
\label{sec:question_decomposition}
We now present evaluation metrics and models for mapping questions into QDMR structures.

\paragraph{Task Definition}
\label{sec:task_definition}
%Our goal is given a NL question $x$ to map it into a sequence of $n$ steps, $\decomplist = \langle s^1, ..., s^n \rangle$ which constitutes its question decomposition. Each step is itself a sequence of tokens, $s^i = (s^{i}_1,...,s^{i}_{m_i})$ which represents a single action that needs to be taken in order to answer the original question (see Table \ref{table:operators}). 

%Recall the definition of QDMR in \S\ref{subsec:definition},
Given a question $x$ we wish to map it to its QDMR steps, $\decomplist = \langle s^1, ..., s^n \rangle$. One can frame this as a sequence-to-sequence problem where $x$ is mapped to a string representing its decomposition. We add a special separating token $\langle\text{SEP}\rangle$, and define the target string to be $s^{1}_1,...,s^{1}_{m_1}, \langle\text{SEP}\rangle, s^{2}_1,...,s^{2}_{m_2}, \langle\text{SEP}\rangle, ...,s^{n}_{m_n}$, where $m_1, ..., m_n$ are the number of tokens in each decomposition step.

\subsection{Evaluation Metrics}
\label{sec:eval_metrics}
We wish to assess the quality of a predicted QDMR, $\hat{\decomplist}$ to a gold standard, $\decomplist$.
Figure~\ref{figure:decomposition_comparison} lists various properties by which question decompositions may differ, such as \emph{granularity} (e.g., steps 1-3 of decomposition 1 are merged into the first step of decomposition 2), \emph{ordering} (e.g., the last two steps are swapped) and \emph{wording} (e.g., using \emph{``from''} instead of \emph{``on''}). While such differences do not affect the overall semantics, the second decomposition can be further decomposed.
To measure such variations, we introduce two types of evaluation metrics. \emph{Sequence-based} metrics treat the decomposition as a sequence of tokens, applying standard text generation metrics. As such metrics ignore the QDMR graph structure, we also employ \emph{graph-based} metrics that compare the predicted graph $G_{\hat{\decomplist}}$ to the gold QDMR graph $G_\decomplist$ (see \S\ref{sec:decomposition_formalism}).

\emph{Sequence-based scores}, where higher values are better, are denoted by $\Uparrow$. \emph{Graph-based scores}, where lower values are better, are denoted by $\Downarrow$.

\begin{itemize}[topsep=0pt, itemsep=0pt, leftmargin=.2in, parsep=0pt]
    \item \textit{Exact Match $\Uparrow$}:
    Measures exact match between $\decomplist$ and $\hat{\decomplist}$, either 0 or 1.
    \item \textit{SARI $\Uparrow$} \cite{xu2016optimizing}:
    \textit{SARI} is commonly used in tasks such as text simplification. Given \decomplist, we consider the sets of added, deleted, and kept n-grams when mapping the question $x$ to $\decomplist$. We compute these three sets for both $\decomplist$ and $\hat{\decomplist}$ using the standard of up to 4-grams, then average (a) the F$_1$ for added n-grams between $\decomplist$ and $\hat{\decomplist}$, (b) the F$_1$ for kept n-grams, and (c) the precision for the deleted n-grams.
 \comment{   
    \item \textit{Match Ratio $\Uparrow$}: 
    Measures the similarity between two token sequences ($\decomplist$, $\hat{\decomplist}$) by aligning identical tokens in the sequences and computing the ratio between successfully aligned tokens and the total length of both strings.

    \item \textit{Structural Match Ratio $\Uparrow$}: 
    Identical to Match Ratio, except we remove all tokens besides $\langle\text{SEP}\rangle$ and reference variables. By matching only these tokens, it measures the structural similarity of QDMRs, namely, the number of steps and the references between steps.
    }
\end{itemize}

\begin{figure}[t]
    \scriptsize
    \centering
    \begin{tabular}{p{7cm}}
         \hline
         \textbf{Question}: \textit{``Show me all the flights from Atlanta to Baltimore on any airline on Thursday''}
         \\ \hline
         \textbf{Decomposition 1}: 
         \begin{enumerate}[nosep,after=\strut]
             \item Return flights 
             \item Return {\color{blue} \#1} from Atlanta 
             \item Return {\color{blue} \#2} to Baltimore 
             \item Return {\color{blue} \#3} on Thursday 
             \item Return {\color{blue} \#4} from any airline
         \end{enumerate}\vspace{-1em}\vspace{-1em}
         \\ \hline
         \textbf{Decomposition 2}: 
         \begin{enumerate}[nosep,after=\strut]
             \item Return flights from Atlanta to Baltimore
             \item Return {\color{blue} \#1} on any airline
             \item Return {\color{blue} \#2} on Thursday 
         \end{enumerate}\vspace{-1em}\vspace{-1em}
         \\ \hline
    \end{tabular}
    \caption{Differences in granularity, step order, and wording between two decompositions.}
    \label{figure:decomposition_comparison}
\end{figure}

\comment{
\begin{table}[t]
    \centering
    \small
    \begin{tabular}{l|c|c}
         Score & Inputs & Range \\
         \hline \hline
         \textit{SARI $\Uparrow$} & \decomplist, $\hat{\decomplist}$ & $[0,1]$ \\
         \textit{Match Ratio $\Uparrow$} & \decomplist, $\hat{\decomplist}$ & $[0,1]$ \\
         \textit{Struct. Match Ratio $\Uparrow$} & \decomplist, $\hat{\decomplist}$ & $[0,1]$ \\
         \hline
         \textit{GED $\Downarrow$} & $G_{\decomplist}$, $G_{\hat{\decomplist}}$ & $[0,1]$ \\
         \textit{Struct. GED $\Downarrow$} & $G_{\decomplist}$, $G_{\hat{\decomplist}}$ & $[0,1]$ \\
         \textit{GED+ $\Downarrow$} & $G_{\decomplist}$, $G_{\hat{\decomplist}}$ & $\mathcal{R}_{\geq 0}$
    \end{tabular}
    \caption{Evaluation scores for measuring similarity and distance between question decompositions. Scores are computed either over pairs of decomposition sequences (\decomplist, $\hat{\decomplist}$) or decomposition graphs ($G_{\decomplist}$, $G_{\hat{\decomplist}}$).}
    \label{table:scores}
\end{table}
}

\begin{figure}[t]
    \centering
    \includegraphics[scale=0.33]{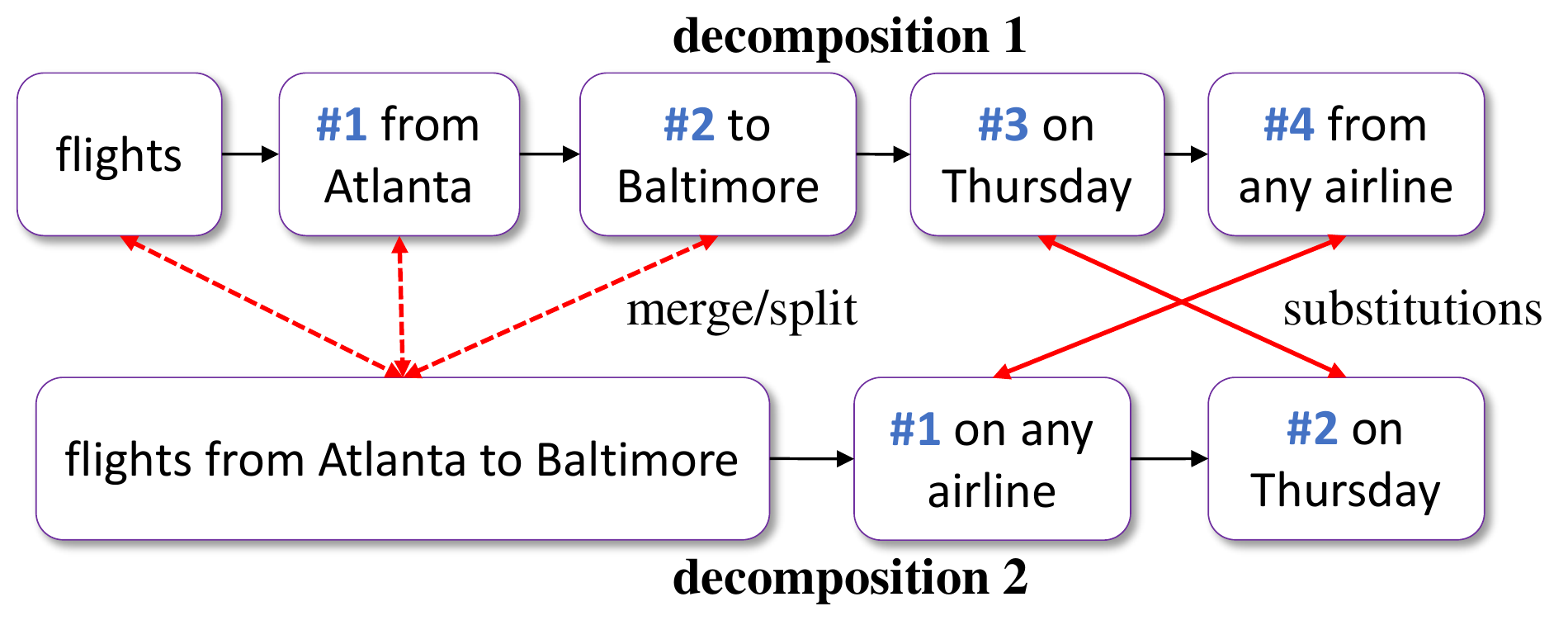}
    \caption{Graph edit operations between the graphs of the two QDMRs in Figure~\ref{figure:decomposition_comparison}.}
    \label{figure:decomposition_correspondence}
\end{figure}

%%Following are the \emph{graph-based} metrics.
\begin{itemize}[topsep=0pt, itemsep=0pt, leftmargin=.2in, parsep=0pt]
    \item \textit{Graph Edit Distance (GED) $\Downarrow$}: 
    A graph edit path is a sequence of node and edge edit operations (addition, deletion, and substitution), where each operation has a predefined cost. \textit{GED} computes the minimal-cost graph edit path required for transitioning from $G_\decomplist$ to $G_{\hat\decomplist}$ (and vice versa), normalized by $\max(|G_\decomplist|, |G_{\hat\decomplist}|)$.
    %We use an approximation algorithm for \textit{GED}.\footnote{\url{https://networkx.github.io/}} 
    Operation costs are $1$ for insertion and deletion of nodes and edges. The substitution cost of two nodes $u,v$ is set to be $1-\textit{Align}(u, v)$, where $\textit{Align}(u, v)$ is the ratio of aligned tokens between these steps. %We normalize \textit{GED} scores by $\max(|G_\decomplist|, |G_{\hat\decomplist}|)$.
    
    \comment{
    \item \textit{Structural GED $\Downarrow$}: Identical to \textit{GED}, but with zero-cost for node substitution. It compares the decomposition structures, i.e., the number of steps (nodes) and references between them (edges), while ignoring their content.
    }
    
    \item \textit{GED+ $\Downarrow$}: 
    Comparing the QDMR graphs in Figure~\ref{figure:decomposition_correspondence}, we consider the splitting and merging of graph nodes. We implement \textit{GED+}, a variant of \textit{GED} with additional operations to merge (split) a set of nodes (node), based on the A* algorithm \cite{hart1968formal}.\footnote{Due to its exponential worst-case complexity, we compute \textit{GED+} only for graphs with up to 5 nodes, covering 75.2\% of the examples in the development set of \textsc{Break}.}
    %Details of the cost function for merge/split are omitted for brevity, but will be part of our public release. It builds on the cost function for node substitution described above, where costs for node addition, deletion and substitution are the same as in \textit{GED}.
    
    %Details on the splitting and merging cost functions are provided in Appendix~\ref{subsec:ged_plus}.
\end{itemize}
%All scores are in the $[0,1]$ interval, except \textit{GED+} which returns a non-negative number.

\begin{table}[t]
\begin{center}
\tiny
\begin{tabular}{p{1.2cm}|p{5.7cm}}\hline
\bf Structure & \bf Example \\ \hline
be-root & How many objects smaller than the matte object \textbf{are} silver \\
& [objects smaller than the matte object, How many \#1 silver] \\ \hline
be-auxpass & Find the average rating star for each movie that \textbf{are} not \textbf{reviewed} by Brittany Harris. \\
& [Brittany Harris, the average rating star for each movie that not reviewed by \#1] \\ \hline
do-subj &  Year \textbf{did} the team with Baltimore Fight Song win the Superbowl? \\
& [team with Baltimore Fight Song, year did \#1 win the Superbowl] \\ \hline
subj-do-have & Which team owned by Malcolm Glazer \textbf{has} Tim Howard playing? \\
& [team Tim Howard playing, \#1 owned by Malcolm Glazer]
 \\ \hline
conjunction & Who trades with China \textbf{and} has a capital city called Khartoum? \\
& [Who has a capital city called Khartoum, \#1 trades with China]
 \\ \hline
how-many & \textbf{How many} metallic objects appear in this image? \\
& [metallic objects appear in this image, the number of \#1]
 \\ \hline
single-prep & Find the ids \textbf{of} the problems reported after 1978. \newline
[the problems reported after 1978, ids of \#1]
 \\ \hline
multi-prep & what flights \textbf{from} Tacoma \textbf{to} Orlando \textbf{on} Saturday \\ & [flights, \#1 from Tacoma, \#2 to Orlando, \#3 on Saturday]  \\ \hline
relcl & Find all the songs \textbf{that} do not have a back vocal. \\
& [all the songs, \#1 that do not have a back vocal]
 \\ \hline
superlative & What is the \textbf{smallest} state bordering ohio \\
& [state bordering ohio, the smallest \#1]
 \\ \hline
acl-verb\comment{*} & Find the first names of students \textbf{studying} in 108. \\
& [students , \#1 studying in 108, first names of \#2]
 \\ \hline
sent-coref & Find \textbf{the claim} that has the largest total settlement amount. Return the effective date of \textbf{the claim}. \\
& [the claim that has the largest total settlement amount, the effective date of \#1] \\ \hline
\end{tabular}
\end{center}
\caption{\label{table:decomposition_rules} The decomposition rules of \rulebased.
  Rules are based on dependency labels, part-of-speech tags and coreference edges. Text fragments used for decomposition are in boldface.
  }
\end{table}

\begin{table*}[t]
\begin{center}
\scriptsize 
\begin{tabular}{l|l|c|c|c|c|c||c}
\hline
 Data & Metric & \copybase{} & \rulebased{} & \seqtoseq{} & \dynamic{} & \copynet{} &  \copynet{} (test) \\ \hline\hline
\multirow{4}{*}{QDMR}& \textit{Exact Match $\Uparrow$} & 0.001 & 0.002 & 0.081 & 0.116 & \bf 0.154 & 0.157\\
%% & \textit{Match Ratio $\Uparrow$} & 0.239 & 0.361 & 0.617 & 0.656 & \bf 0.681 & 0.668\\
%% & \textit{Struct. Match Ratio $\Uparrow$} & 0.003 & 0.378 & 0.803 & 0.820 & \bf 0.821 & 0.818\\ 
& \textit{SARI $\Uparrow$}  & 0.431 & 0.508 & 0.665 & 0.705 & \bf 0.748 & 0.746\\ 
& \textit{GED $\Downarrow$}  & 0.937 & 0.799 & 0.398 & 0.363 & \bf 0.318 & 0.322\\ 
%% & \textit{Struct. GED $\Downarrow$}  & 0.832 & 0.593 & 0.217 & \bf 0.189 & 0.194 & 0.192\\ 
& \textit{GED+ $\Downarrow$}  & 1.813 & 1.722 & 1.424 & 1.137 & \bf 0.941 & 0.984\\ \hline
\multirow{4}{*}{QDMR$_{\textit{high}}$}& \textit{Exact Match $\Uparrow$} & 0.001 & 0.010 & 0.001 & 0.015 & \bf 0.081 & 0.083\\
%% & \textit{Match Ratio $\Uparrow$} & 0.466 & 0.476 & 0.295 & 0.540 & \bf 0.699 & 0.708\\ 
%% & \textit{Struct. Match Ratio $\Uparrow$}  & 0.109 & 0.482 & 0.752 & 0.718 & \bf 0.779 & 0.792\\ 
& \textit{SARI $\Uparrow$}  & 0.501 & 0.554 & 0.379 & 0.504 & \bf 0.722 & 0.722\\ 
& \textit{GED $\Downarrow$}  & 0.793 & 0.659 & 0.585 & 0.468 & \bf 0.319 & 0.316\\ 
%% & \textit{Struct. GED $\Downarrow$}  & 0.662 & 0.457 & 0.236 & 0.233 & \bf 0.183 & 0.169\\ 
& \textit{GED+ $\Downarrow$}  & 1.102 & 1.395 & 1.655 & 1.238 & \bf 0.716 & 0.709 \\ \hline  \end{tabular}
\end{center}
\caption{\label{table:experiments} Performance of QDMR parsing models on the development and test set.
  \textit{GED+} is computed only for the subset of QDMR graphs with up to 5 nodes, covering 66.1\% of QDMRs and 97.6\% of high-level data.}
\end{table*}

\subsection{QDMR Parsing Models}
\label{subsec:baselines}

We present models for QDMR parsing, built over AllenNLP \cite{Gardner2017AllenNLP}.
\begin{itemize}[topsep=4pt, itemsep=1pt, leftmargin=.2in, parsep=2pt]
    \item \copybase{}: A model that copies the input question $x$, without introducing any modifications. 
    \item \rulebased{}: We defined 12 decomposition rules, to be applied over the dependency tree of the question, augmented with coreference relations.
    %JB: for brevity - we can add back for camera-ready.
    %\footnote{We use spaCy (\url{https://spacy.io}) and NeuralCoref (\url{https://github.com/huggingface/neuralcoref})} 
    A rule is a regular expression over the question dependency tree, which invokes a decomposition operation when matched (Table~\ref{table:decomposition_rules}). E.g., the rule for relative clauses (\texttt{relcl}) breaks the question at the relative pronoun \emph{``that''}, while adding a reference to the preceding part of the sentence.
    A full decomposition is obtained by recursively applying the rules until no rule is matched.
    \item \seqtoseq{}: A sequence-to-sequence neural model with a 5-layer LSTM encoder and attention at decoding time.
    
    % TW - removed for lack of space
    
    \item \dynamic{}: \seqtoseq{} with a dynamic output vocabulary restricted to the closed set of tokens $L_x$ available to crowd-workers (see \S\ref{sec:data_collection}).

    \item \copynet{}: \seqtoseq{} with an added copy mechanism that allows copying tokens from the input sequence \cite{gu2016copying}. 
\end{itemize}

%JB: no space I think.
%Further details on the network architectures and training procedure are provided in Appendix~\ref{subsec:supplemental_neural}. 

%Comparing models trained on \textsc{Break} against a rule-based decomposition is done to evaluate how much of question decomposition can be captured by simple syntactic rules.

\subsection{Results} 
\label{sec:experiments_results}

Table \ref{table:experiments} presents model performance on \datasetname{}. 
Neural models outperform the \rulebased{} baseline and perform reasonably well, with \copynet{} obtaining the best scores across all metrics. This can be attributed to most of the tokens in a QDMR parse being copied from the original question.
%Overall, the best performing model appears to be \copynet{}. Its advantage over \seqtoseq{} in can be explained given decompositions consist mainly of tokens from the original question, enabling \copynet{} to generalize better during training.

% TW - incorporated most of this with the text above.
\comment{Not performing any decomposition steps, \copybase{} obtained the lowest scores among all baselines, as expected.
The \rulebased{} model falls short of achieving comparable results to the models trained over \textsc{Break}.
\tw{Should refer to \dynamic{} and its performance.}
The \copynet{} model achieves the best results across all the evaluation metrics. The advantage of \copynet{} over \seqtoseq{} in this task can by explained by the fact that a question decomposition mostly consists of tokens from the question, making the training process faster for \copynet{}. Yet, there is a still a room for improvement, as the optimal scores are 1 for the string based metrics and 0 for the graph based metrics.}

\paragraph{Error analysis}\label{subsec:error_analysis}
To judge the quality of predicted QDMRs we sampled 100 predictions of \copynet{} (Table \ref{table:analysis}) half of them being \textit{high-level} QDMRs. 
%Our qualitative analysis shows that as questions are more structured the model performance is better. 
For standard QDMR, 24\% of the sampled predictions were an exact match, with an additional 30\% being fully decomposed and semantically equivalent to the gold decompositions. E.g., in the first row of Table \ref{table:analysis}, the gold decomposition first discards the number of cylinders, then counts the remaining objects. Instead, \copynet{} opted to count both groups, then subtract the number of cylinders from the number of objects. This illustrates how different QDMRs may be equivalent. 

For high-level examples (from RC datasets), as questions are often less structured, they require a deeper semantic understating from the decomposition model. Only 8\% of the predictions were an exact match, with an additional 46\% being semantically equivalent to the gold. The remaining 46\%  were of erroneous predictions (see Table \ref{table:analysis}). 
%We note that graph-based evaluation metrics might prove less indicative of \textit{high-level} QDMR quality, as they are shorter and have lower operator diversity (Tables \ref{table:operator_prevalence}-\ref{table:qdmr_lengths}) compared to granular QDMR.

\comment{
\paragraph{Domain-independent Decomposition}\label{subsec:experiments_modality}
A motivation for developing QDMR was for it to serve as a general-purpose decomposition formalism. As \datasetname{} contains questions from multiple sources (images, text, KBs), we sought to test whether data augmentation over a different modalities could improve QDMR decomposition. For our experiments we sampled 16K examples from \datasetname{}, half over KBs \cite{Yu2018SpiderAL,johnson2017clevr} and half over text \cite{talmor2018web, dua2019drop}. Two \copynet{} models were trained: (1) over the 8K KB examples; \jb{which 8K? there are two, maybe try both?} (2) over all 16K examples on both structured and unstructured sources.
Results in Table \ref{} show the model trained on both KB and text examples outperforms the first. This further establishes the shared structure between questions across different modalities. \jb{still need to do this and write slightly more about this}
}

\section{Related Work}

\begin{table*}[t]
\begin{center}
\tiny
\begin{tabular}{p{4.1cm}|p{4.2cm}|p{4.2cm}|p{1cm}}
 \hline
 \bf Question & \bf Gold & \bf Prediction (\copynet{}) & \bf Analysis\\ \hline\hline
   \textit{``How many objects other than cylinders are there?''} & (1) objects; (2) cylinders; (3) {\color{blue}\#1} besides {\color{blue}\#2}; (4) number of {\color{blue}\#3}. & (1) objects; (2) cylinders; (3) number of {\color{blue}\#1}; (4) number of {\color{blue}\#2}; (5) difference of {\color{blue}\#3} and {\color{blue}\#4}. & sem. equiv. \bf(30\%) \\ \hline
   \textit{``Where is the youngest teacher from?''} & (1) teachers; (2) the youngest of {\color{blue}\#1}; (3) where is {\color{blue}\#2} from. & (1) youngest teacher; (2) where is {\color{blue}\#1}. &  incorrect \bf(46\%) \\ \hline\hline
   \textit{``Kyle York is the Chief Strategy Officer of a company acquired by what corporation in 2016?''} & (1) company that Kyle York is the Chief Strategy Officer of; (2) corporation that acquired {\color{blue}\#1} in 2016. & (1) company that Kyle York is the Chief Strategy Officer of; (2) corporation in 2016 that {\color{blue}\#1} was acquired by.  & sem. equiv. \bf (46\%)  \\ \hline
   \textit{``Dayton's Devils had a cameo from the `MASH' star who played what role on the show?''} & (1) MASH star that Dayton 's Devils had a cameo from; (2) role that {\color{blue}\#1} played on the show. & (1) the MASH that Dayton 's Devils had a cameo; (2) what role on the show star of {\color{blue}\#1} played. & incorrect \bf (46\%) \\ \hline
\end{tabular}
\end{center}
\caption{\label{table:analysis} Manual error analysis of the \copynet{} model predictions. Lower examples are of \textit{high-level} QDMRs. }
\end{table*}

\paragraph{Question decomposition}
% Question Decomposition 
% JB: we said this already
%Increasing work has been dedicated to the creation of compositional questions' datasets that require reasoning \cite{johnson2017clevr, welbl2017constructing, yang2018HotpotQAAD}.
Recent work on QA through question decomposition has focused mostly on single modalities \cite{Gupta2018NeuralCD, GuoIRNet2019, min2019multi}. QA using neural modular networks has been suggested for both KBs and images by \newcite{andreas2016learning} and \newcite{hu2017learning}. Question decomposition over text was proposed by \citet{talmor2018web}, however over a much more limited set of questions than in \datasetname{}.
\newcite{iyyer2017search} have also decomposed questions to create a ``sequential question answering'' task. Their annotators viewed a web table and performed actions over it to retrieve the cells that constituted the answer.
Conversely, we provided annotators only with the question, as QDMR is agnostic to the original context.

An opposite annotation cycle to ours was presented in \citet{Cheng2018BuildingAN}. The authors generate sequences of simple questions which crowd-workers paraphrase into a compositional question. Questions in \datasetname{} are composed by humans, and are then decomposed to QDMR.

% TW - Removed for lack of space

 %Related to question decomposition, \citet{zhang-etal-2018-cross} proposed a formalism aimed at ``cross-lingual semantic parsing'', mapping sentences in a source language (e.g., Chinese) to a universal decompositional semantic analysis.
 
 %The work of \citet{narayan2017split} can be equated to QDMR parsing in terms of sentence split and simplification. In contrast, QDMRs includes reasoning operators that are absent from the original question but necessary for its solution.

% Sequential question answering

%\paragraph{Sequential Question Answering (SQA)}
%SQA \cite{long2016projections, iyyer2017search, suhr-etal-2018-learning} is concerned with mapping context dependent questions to structured queries. The sequence of context dependent questions often expresses a single intent. Though this sequence can be regarded as a question decomposition of the intent, it differs from QDMR as steps are freely expressed, or include further exploratory sub-questions that were not expressed in the original intent \cite{Yu2019SParCAL}. 

\paragraph{Semantic formalism annotation}
Labeling corpora with a semantic formalism has often been reserved for expert annotators \cite{dahl1994expanding, zelle96geoquery, Abend2013UniversalCC, Yu2018SpiderAL}. Recent work has focused on cheaply eliciting quality annotations from non-experts through crowdsourcing \cite{he2016human, iyer2017neural, michael2018qamr}. \citet{FitzGerald2018LargeScaleQP} facilitated non-expert annotation by introducing a formalism expressed in natural language for semantic-role-labeling. This mirrors QDMR, as both are expressed in natural language.

\paragraph{Relation to other formalisms}
%DCS
QDMR is related to Dependency-based Compositional Semantics \cite{liang13cl}, as both focus on question representations. However, QDMR is designed to facilitate annotations, while DCS is centered on paralleling syntax.
%Closely related to our work is the DCS meaning representation for questions \cite{liang13cl}, designed to parallel syntactic dependency trees to facilitate parsing. While both are semantic parsing oriented, QDMR is designed to facilitate annotations, being expressed through phrases rather than DCS trees. 
%CCG
Domain-independent intermediate representations for semantic parsers were proposed by \citet{kwiatkowski2013scaling} and \citet{reddy2016transforming}.
As there is no consensus on the ideal meaning representation for semantic parsing, representations are often chosen based on the particular execution setup: SQL is used for relational databases \cite{Yu2018SpiderAL}, SPARQL for graph KBs \cite{yih2016value}, while other ad-hoc languages are used based on the task at hand.
%\citet{GuoIRNet2019} proposed SemQL as a  variant of SQL that improves supervised learning. 
We frame QDMR as an easy-to-annotate formalism that can be potentially converted to other representations, depending on the task. 
%AMR
Last, AMR \cite{banarescu2013amr} is a meaning representation for sentences. Instead of representing general language, QDMR represents questions, which are important for QA systems, and for probing models for reasoning.

\section{Conclusion}

In this paper, we presented a formalism for question understanding.
We have shown it is possible to train crowd-workers to produce such representations with high quality at scale, and created \datasetname{}, a benchmark for question decomposition with over 83K decompositions of questions from 10 datasets and 3 modalities (DB, images, text).  
We presented the utility of QDMR for both open-domain question answering and semantic parsing, and constructed a QDMR parser with reasonable performance.
%In future work we intend to develop better parsers by taking into account the structure of QDMR more explicitly, and to investigate the utility of QDMR for textual and visual reasoning. 
QDMR proposes a promising direction for modeling question understanding, which we believe will be useful for multiple tasks in which reasoning is probed through questions.

\section*{Acknowledgments}
This work was completed in partial fulfillment for the PhD of Tomer Wolfson.
This research was partially supported by The Israel Science Foundation grants 942/16 and 978/17, The Yandex Initiative for Machine Learning and the European Research Council (ERC) under the European Union Horizons 2020 research and innovation programme (grant ERC DELPHI 802800).

\bibliography{all}

\begin{thebibliography}{53}
\expandafter\ifx\csname natexlab\endcsname\relax\def\natexlab#1{#1}\fi

\bibitem[{Abend and Rappoport(2013)}]{Abend2013UniversalCC}
Omri Abend and Ari Rappoport. 2013.
\newblock Universal conceptual cognitive annotation (ucca).
\newblock In \emph{Association for Computational Linguistics (ACL)}.

\bibitem[{Abujabal et~al.(2019)Abujabal, Roy, Yahya, and
  Weikum}]{Abujabal2018ComQAAC}
Abdalghani Abujabal, Rishiraj~Saha Roy, Mohamed Yahya, and Gerhard Weikum.
  2019.
\newblock Comqa: A community-sourced dataset for complex factoid question
  answering with paraphrase clusters.
\newblock In \emph{North American Association for Computational Linguistics
  (NAACL)}.

\bibitem[{Andreas et~al.(2016)Andreas, Rohrbach, Darrell, and
  Klein}]{andreas2016learning}
Jacob Andreas, Marcus Rohrbach, Trevor Darrell, and Dan Klein. 2016.
\newblock Learning to compose neural networks for question answering.
\newblock In \emph{Human Language Technology and North American Association for
  Computational Linguistics (HLT/NAACL)}.

\bibitem[{Androutsopoulos et~al.(1995)Androutsopoulos, Ritchie, and
  Thanisch}]{androutsopoulos95nlidb}
Ion Androutsopoulos, Graeme~D. Ritchie, and Peter Thanisch. 1995.
\newblock Natural language interfaces to databases -- an introduction.
\newblock \emph{Journal of Natural Language Engineering}, 1:29--81.

\bibitem[{Antol et~al.(2015)Antol, Agrawal, Lu, Mitchell, Batra, Zitnick, and
  Parikh}]{antol2015vqa}
Stanislaw Antol, Aishwarya Agrawal, Jiasen Lu, Margaret Mitchell, Dhruv Batra,
  C.~Lawrence Zitnick, and Devi Parikh. 2015.
\newblock Vqa: Visual question answering.
\newblock In \emph{International Conference on Computer Vision (ICCV)}, pages
  2425--2433.

\bibitem[{Baik et~al.(2019)Baik, Jagadish, and Li}]{baik2019bridging}
Christopher Baik, Hosagrahar~Visvesvaraya Jagadish, and Yunyao Li. 2019.
\newblock Bridging the semantic gap with sql query logs in natural language
  interfaces to databases.
\newblock \emph{2019 IEEE 35th International Conference on Data Engineering
  (ICDE)}, pages 374--385.

\bibitem[{Banarescu et~al.(2013)Banarescu, Bonial, Cai, Georgescu, Griffitt,
  Hermjakob, Knight, Koehn, Palmer, and Schneider}]{banarescu2013amr}
Laura Banarescu, Claire Bonial, Shu Cai, Madalina Georgescu, Kira Griffitt, Ulf
  Hermjakob, Kevin Knight, Philipp Koehn, Martha Palmer, and Nathan Schneider.
  2013.
\newblock Abstract meaning representation for sembanking.
\newblock In \emph{7th Linguistic Annotation Workshop and Interoperability with
  Discourse}.

\bibitem[{Chamberlin and Boyce(1974)}]{chamberlin1974sequel}
Donald~D Chamberlin and Raymond~F Boyce. 1974.
\newblock Sequel: A structured english query language.
\newblock In \emph{Proceedings of the 1974 ACM SIGFIDET (now SIGMOD) workshop
  on Data description, access and control}, pages 249--264. ACM.

\bibitem[{Chen et~al.(2017)Chen, Fisch, Weston, and Bordes}]{chen2017reading}
Danqi Chen, Adam Fisch, Jason Weston, and Antoine Bordes. 2017.
\newblock Reading {W}ikipedia to answer open-domain questions.
\newblock In \emph{Association for Computational Linguistics (ACL)}.

\bibitem[{Chen and Durrett(2019)}]{Chen2019UnderstandingDD}
Jifan Chen and Greg Durrett. 2019.
\newblock Understanding dataset design choices for multi-hop reasoning.
\newblock In \emph{Association for Computational Linguistics (ACL)}.

\bibitem[{Cheng et~al.(2018)Cheng, Reddy, and Lapata}]{Cheng2018BuildingAN}
Jianpeng Cheng, Siva Reddy, and Mirella Lapata. 2018.
\newblock Building a neural semantic parser from a domain ontology.
\newblock \emph{ArXiv}, abs/1812.10037.

\bibitem[{Choi et~al.(2015)Choi, Kwiatkowski, and
  Zettlemoyer}]{choi-etal-2015-scalable}
Eunsol Choi, Tom Kwiatkowski, and Luke Zettlemoyer. 2015.
\newblock Scalable semantic parsing with partial ontologies.
\newblock In \emph{Association for Computational Linguistics (ACL)}.

\bibitem[{Clarke et~al.(2010)Clarke, Goldwasser, Chang, and
  Roth}]{clarke10world}
James Clarke, Dan Goldwasser, Ming-Wei Chang, and Dan Roth. 2010.
\newblock Driving semantic parsing from the world's response.
\newblock In \emph{Computational Natural Language Learning (CoNLL)}, pages
  18--27.

\bibitem[{Codd(1970)}]{codd1970relational}
Edgar~F Codd. 1970.
\newblock A relational model of data for large shared data banks.
\newblock \emph{Communications of the ACM}, 13(6):377--387.

\bibitem[{Dahl et~al.(1994)Dahl, Bates, Brown, Fisher, Hunicke-Smith, Pallett,
  Pao, Rudnicky, and Shriberg}]{dahl1994expanding}
Deborah~A. Dahl, Madeleine Bates, Michael Brown, William~M. Fisher, Kate
  Hunicke-Smith, David~S. Pallett, Christine Pao, Alexander~I. Rudnicky, and
  Elizabeth Shriberg. 1994.
\newblock Expanding the scope of the {ATIS} task: The {ATIS-3} corpus.
\newblock In \emph{Workshop on Human Language Technology}, pages 43--48.

\bibitem[{Devlin et~al.(2019)Devlin, Chang, Lee, and
  Toutanova}]{devlin2018bert}
Jacob Devlin, Ming-Wei Chang, Kenton Lee, and Kristina Toutanova. 2019.
\newblock Bert: Pre-training of deep bidirectional transformers for language
  understanding.
\newblock In \emph{North American Association for Computational Linguistics
  (NAACL)}.

\bibitem[{Dua et~al.(2019)Dua, Wang, Dasigi, Stanovsky, Singh, and
  Gardner}]{dua2019drop}
Dheeru Dua, Yizhong Wang, Pradeep Dasigi, Gabriel Stanovsky, Sameer Singh, and
  Matt Gardner. 2019.
\newblock {DROP}: A reading comprehension benchmark requiring discrete
  reasoning over paragraphs.
\newblock In \emph{Human Language Technology and North American Association for
  Computational Linguistics (HLT/NAACL)}.

\bibitem[{FitzGerald et~al.(2018)FitzGerald, Michael, He, and
  Zettlemoyer}]{FitzGerald2018LargeScaleQP}
Nicholas FitzGerald, Julian Michael, Luheng He, and Luke~S. Zettlemoyer. 2018.
\newblock Large-scale qa-srl parsing.
\newblock In \emph{Association for Computational Linguistics (ACL)}.

\bibitem[{Gardner et~al.(2017)Gardner, Grus, Neumann, Tafjord, Dasigi, Liu,
  Peters, Schmitz, and Zettlemoyer}]{Gardner2017AllenNLP}
Matt Gardner, Joel Grus, Mark Neumann, Oyvind Tafjord, Pradeep Dasigi,
  Nelson~F. Liu, Matthew Peters, Michael Schmitz, and Luke~S. Zettlemoyer.
  2017.
\newblock \href {http://arxiv.org/abs/arXiv:1803.07640} {Allennlp: A deep
  semantic natural language processing platform}.

\bibitem[{Gu et~al.(2016)Gu, Lu, Li, and Li}]{gu2016copying}
Jiatao Gu, Zhengdong Lu, Hang Li, and Victor O.~K. Li. 2016.
\newblock Incorporating copying mechanism in sequence-to-sequence learning.
\newblock In \emph{Association for Computational Linguistics (ACL)}.

\bibitem[{Guo et~al.(2019)Guo, Zhan, Gao, Xiao, Lou, Liu, and
  Zhang}]{GuoIRNet2019}
Jiaqi Guo, Zecheng Zhan, Yan Gao, Yan Xiao, Jian-Guang Lou, Ting Liu, and
  Dongmei Zhang. 2019.
\newblock Towards complex text-to-sql in cross-domain database with
  intermediate representation.
\newblock In \emph{Association for Computational Linguistics (ACL)}.

\bibitem[{Gupta and Lewis(2018)}]{Gupta2018NeuralCD}
Nitish Gupta and Mike Lewis. 2018.
\newblock Neural compositional denotational semantics for question answering.
\newblock In \emph{Empirical Methods in Natural Language Processing (EMNLP)}.

\bibitem[{Hart et~al.(1968)Hart, Nilsson, and Raphael}]{hart1968formal}
Peter~E. Hart, Nils~J. Nilsson, and Bertram Raphael. 1968.
\newblock A formal basis for the heuristic determination of minimum cost paths.
\newblock \emph{IEEE transactions on Systems Science and Cybernetics},
  4(2):100--107.

\bibitem[{He et~al.(2016)He, Michael, Lewis, and Zettlemoyer}]{he2016human}
Luheng He, Julian Michael, Mike Lewis, and Luke Zettlemoyer. 2016.
\newblock Human-in-the-loop parsing.
\newblock In \emph{Empirical Methods in Natural Language Processing (EMNLP)}.

\bibitem[{Hu et~al.(2017)Hu, Andreas, Rohrbach, Darrell, and
  Saenko}]{hu2017learning}
Ronghang Hu, Jacob Andreas, Marcus Rohrbach, Trevor Darrell, and Kate Saenko.
  2017.
\newblock Learning to reason: End-to-end module networks for visual question
  answering.
\newblock In \emph{International Conference on Computer Vision (ICCV)}.

\bibitem[{Hudson and Manning(2019)}]{Hudson_2019_CVPR}
Drew~A. Hudson and Christopher~D. Manning. 2019.
\newblock Gqa: A new dataset for real-world visual reasoning and compositional
  question answering.
\newblock In \emph{The IEEE Conference on Computer Vision and Pattern
  Recognition (CVPR)}.

\bibitem[{Iyer et~al.(2017)Iyer, Konstas, Cheung, Krishnamurthy, and
  Zettlemoyer}]{iyer2017neural}
Srini Iyer, Ioannis Konstas, Alvin Cheung, Jayant Krishnamurthy, and Luke
  Zettlemoyer. 2017.
\newblock Learning a neural semantic parser from user feedback.
\newblock In \emph{Association for Computational Linguistics (ACL)}.

\bibitem[{Iyyer et~al.(2017)Iyyer, Yih, and Chang}]{iyyer2017search}
Mohit Iyyer, Wen-tau Yih, and Ming-Wei Chang. 2017.
\newblock Search-based neural structured learning for sequential question
  answering.
\newblock In \emph{Proceedings of the 55th Annual Meeting of the Association
  for Computational Linguistics (Volume 1: Long Papers)}, pages 1821--1831.

\bibitem[{Jiang and Bansal(2019)}]{jiang-bansal-2019-avoiding}
Yichen Jiang and Mohit Bansal. 2019.
\newblock Avoiding reasoning shortcuts: Adversarial evaluation, training, and
  model development for multi-hop {QA}.
\newblock In \emph{Association for Computational Linguistics (ACL)}.

\bibitem[{Johnson et~al.(2017)Johnson, Hariharan, van~der Maaten, Fei-Fei,
  Zitnick, and Girshick}]{johnson2017clevr}
Justin Johnson, Bharath Hariharan, Laurens van~der Maaten, Li~Fei-Fei,
  C.~Lawrence Zitnick, and Ross~B. Girshick. 2017.
\newblock Clevr: A diagnostic dataset for compositional language and elementary
  visual reasoning.
\newblock In \emph{Computer Vision and Pattern Recognition (CVPR)}.

\bibitem[{Kwiatkowski et~al.(2013)Kwiatkowski, Choi, Artzi, and
  Zettlemoyer}]{kwiatkowski2013scaling}
Tom Kwiatkowski, Eunsol Choi, Yoav Artzi, and Luke Zettlemoyer. 2013.
\newblock Scaling semantic parsers with on-the-fly ontology matching.
\newblock In \emph{Empirical Methods in Natural Language Processing (EMNLP)}.

\bibitem[{Kwiatkowski et~al.(2019)Kwiatkowski, Palomaki, Redfield, Collins,
  Parikh, Alberti, Epstein, Polosukhin, Devlin, Lee
  et~al.}]{kwiatkowski2019natural}
Tom Kwiatkowski, Jennimaria Palomaki, Olivia Redfield, Michael Collins, Ankur
  Parikh, Chris Alberti, Danielle Epstein, Illia Polosukhin, Jacob Devlin,
  Kenton Lee, et~al. 2019.
\newblock Natural questions: a benchmark for question answering research.
\newblock \emph{Transactions of the Association for Computational Linguistics},
  7:453--466.

\bibitem[{Li and Jagadish(2014)}]{Li2014NaLIRAI}
Fei Li and Hosagrahar~Visvesvaraya Jagadish. 2014.
\newblock Nalir: an interactive natural language interface for querying
  relational databases.
\newblock In \emph{International Conference on Management of Data, {SIGMOD}}.

\bibitem[{Li et~al.(2014)Li, Pan, and Jagadish}]{li2014schema}
Fei Li, Tianyin Pan, and Hosagrahar~Visvesvaraya Jagadish. 2014.
\newblock Schema-free {SQL}.
\newblock In \emph{International Conference on Management of Data, {SIGMOD}},
  pages 1051--1062.

\bibitem[{Liang et~al.(2013)Liang, Jordan, and Klein}]{liang13cl}
Percy Liang, Michael~I. Jordan, and Dan Klein. 2013.
\newblock Learning dependency-based compositional semantics.
\newblock \emph{Computational Linguistics}, 39:389--446.

\bibitem[{Michael et~al.(2018)Michael, Stanovsky, He, Dagan, and
  Zettlemoyer}]{michael2018qamr}
Julian Michael, Gabriel Stanovsky, Luheng He, Ido Dagan, and Luke Zettlemoyer.
  2018.
\newblock Crowdsourcing question--answer meaning representations.
\newblock In \emph{North American Association for Computational Linguistics
  (NAACL)}.

\bibitem[{Min et~al.(2019{\natexlab{a}})Min, Wallace, Singh, Gardner,
  Hajishirzi, and Zettlemoyer}]{min2019compositional}
Sewon Min, Eric Wallace, Sameer Singh, Matt Gardner, Hannaneh Hajishirzi, and
  Luke Zettlemoyer. 2019{\natexlab{a}}.
\newblock Compositional questions do not necessitate multi-hop reasoning.
\newblock In \emph{Association for Computational Linguistics (ACL)}.

\bibitem[{Min et~al.(2019{\natexlab{b}})Min, Zhong, Zettlemoyer, and
  Hajishirzi}]{min2019multi}
Sewon Min, Victor Zhong, Luke Zettlemoyer, and Hannaneh Hajishirzi.
  2019{\natexlab{b}}.
\newblock Multi-hop reading comprehension through question decomposition and
  rescoring.
\newblock In \emph{Association for Computational Linguistics (ACL)}.

\bibitem[{Pasupat and Liang(2015)}]{pasupat2015compositional}
Panupong Pasupat and Percy Liang. 2015.
\newblock Compositional semantic parsing on semi-structured tables.
\newblock In \emph{Association for Computational Linguistics (ACL)}.

\bibitem[{Pelletier(1994)}]{Pelletier1994}
Francis~Jeffry Pelletier. 1994.
\newblock \href {https://doi.org/10.1007/BF00763644} {The principle of semantic
  compositionality}.
\newblock \emph{Topoi}, 13(1):11--24.

\bibitem[{Price(1990)}]{price1990atis}
P.~J. Price. 1990.
\newblock Evaluation of spoken language systems: The {ATIS} domain.
\newblock In \emph{Proceedings of the Third DARPA Speech and Natural Language
  Workshop}, pages 91--95.

\bibitem[{Qi et~al.(2019)Qi, Lin, Mehr, Wang, and
  Manning}]{qi-etal-2019-answering}
Peng Qi, Xiaowen Lin, Leo Mehr, Zijian Wang, and Christopher~D. Manning. 2019.
\newblock Answering complex open-domain questions through iterative query
  generation.
\newblock In \emph{Empirical Methods in Natural Language Processing (EMNLP)},
  pages 2590--2602. Association for Computational Linguistics.

\bibitem[{Rajpurkar et~al.(2016)Rajpurkar, Zhang, Lopyrev, and
  Liang}]{rajpurkar2016squad}
Pranav Rajpurkar, Jian Zhang, Konstantin Lopyrev, and Percy Liang. 2016.
\newblock {SQuAD}: 100,000+ questions for machine comprehension of text.
\newblock In \emph{Empirical Methods in Natural Language Processing (EMNLP)}.

\bibitem[{Reddy et~al.(2016)Reddy, T{\"a}ckstr{\"o}m, Collins, Kwiatkowski,
  Das, Steedman, and Lapata}]{reddy2016transforming}
Siva Reddy, Oscar T{\"a}ckstr{\"o}m, Michael Collins, Tom Kwiatkowski, Dipanjan
  Das, Mark Steedman, and Mirella Lapata. 2016.
\newblock Transforming dependency structures to logical forms for semantic
  parsing.
\newblock In \emph{Association for Computational Linguistics (ACL)}.

\bibitem[{Suhr et~al.(2019)Suhr, Zhou, Zhang, Bai, and Artzi}]{suhr2018corpus}
Alane Suhr, Stephanie Zhou, Iris Zhang, Huajun Bai, and Yoav Artzi. 2019.
\newblock A corpus for reasoning about natural language grounded in
  photographs.
\newblock \emph{Association for Computational Linguistics (ACL)}.

\bibitem[{Talmor and Berant(2018)}]{talmor2018web}
Alon Talmor and Jonathan Berant. 2018.
\newblock The web as knowledge-base for answering complex questions.
\newblock In \emph{North American Association for Computational Linguistics
  (NAACL)}.

\bibitem[{Welbl et~al.(2018)Welbl, Stenetorp, and
  Riedel}]{welbl2017constructing}
Johannes Welbl, Pontus Stenetorp, and Sebastian Riedel. 2018.
\newblock Constructing datasets for multi-hop reading comprehension across
  documents.
\newblock \emph{Transactions of the Association for Computational Linguistics},
  6:287--302.

\bibitem[{Xu et~al.(2016)Xu, Napoles, Pavlick, Chen, and
  Callison-Burch}]{xu2016optimizing}
Wei Xu, Courtney Napoles, Ellie Pavlick, Quanze Chen, and Chris Callison-Burch.
  2016.
\newblock Optimizing statistical machine translation for text simplification.
\newblock \emph{Transactions of the Association for Computational Linguistics},
  4:401--415.

\bibitem[{Yang et~al.(2018)Yang, Qi, Zhang, Bengio, Cohen, Salakhutdinov, and
  Manning}]{yang2018HotpotQAAD}
Zhilin Yang, Peng Qi, Saizheng Zhang, Yoshua Bengio, William~W. Cohen,
  Ruslan~R. Salakhutdinov, and Christopher~D. Manning. 2018.
\newblock Hotpotqa: A dataset for diverse, explainable multi-hop question
  answering.
\newblock In \emph{Empirical Methods in Natural Language Processing (EMNLP)}.

\bibitem[{Yih et~al.(2016)Yih, Richardson, Meek, Chang, and Suh}]{yih2016value}
Wen{-}tau Yih, Matthew Richardson, Christopher Meek, Ming-Wei Chang, and Jina
  Suh. 2016.
\newblock The value of semantic parse labeling for knowledge base question
  answering.
\newblock In \emph{Association for Computational Linguistics (ACL)}.

\bibitem[{Yu et~al.(2018)Yu, Zhang, Yang, Yasunaga, Wang, Li, Ma, Li, Yao,
  Roman, Zhang, and Radev}]{Yu2018SpiderAL}
Tao Yu, Rui Zhang, Kai Yang, Michihiro Yasunaga, Dongxu Wang, Zifan Li, James
  Ma, Irene Li, Qingning Yao, Shanelle Roman, Zilin Zhang, and Dragomir~R.
  Radev. 2018.
\newblock Spider: A large-scale human-labeled dataset for complex and
  cross-domain semantic parsing and text-to-sql task.
\newblock In \emph{Empirical Methods in Natural Language Processing (EMNLP)}.

\bibitem[{Zelle and Mooney(1996)}]{zelle96geoquery}
John~M. Zelle and Raymond~J. Mooney. 1996.
\newblock Learning to parse database queries using inductive logic programming.
\newblock In \emph{Association for the Advancement of Artificial Intelligence
  (AAAI)}, pages 1050--1055.

\bibitem[{Zettlemoyer and Collins(2005)}]{zettlemoyer05ccg}
Luke Zettlemoyer and Michael Collins. 2005.
\newblock Learning to map sentences to logical form: Structured classification
  with probabilistic categorial grammars.
\newblock In \emph{Uncertainty in Artificial Intelligence (UAI)}, pages
  658--666.

\end{thebibliography}
\bibliographystyle{acl_natbib}

% TW - remove appendix from submission
\comment{
\newpage
\appendix

\section{Supplementary Material}
\label{sec:supplemental}

\subsection{QDMR Annotation Lexicon}
\label{subsec:supplemental_annotation_lexicon}
In annotating QDMR steps we present workers with a closed lexicon of tokens. These include question and stop words: \texttt{a}, \texttt{if}, \texttt{how}, \texttt{where}, \texttt{when}, \texttt{which}, \texttt{who}, \texttt{what}, \texttt{with}, \texttt{was}, \texttt{did}, \texttt{to}, \texttt{from}, \texttt{both}, \texttt{and}, \texttt{or}, \texttt{the}, \texttt{of}, \texttt{is}, \texttt{are}, \texttt{besides}, \texttt{that}, \texttt{have}, \texttt{has}, \texttt{for each}, \texttt{number of}, \texttt{not}, \texttt{than}, \texttt{those}, \texttt{on}, \texttt{in}, \texttt{any}, \texttt{there}, \texttt{,}.

Comparison and arithmetic tokens: \texttt{same as}, \texttt{higher than}, \texttt{larger than}, \texttt{smaller than}, \texttt{lower than}, \texttt{more}, \texttt{less}, \texttt{at least}, \texttt{at most}, \texttt{equal}, \texttt{highest}, \texttt{lowest}, \texttt{sorted by}, \texttt{sum}, \texttt{difference}, \texttt{multiplication}, \texttt{division}, \texttt{100}, \texttt{hundred}, \texttt{one}, \texttt{two}, \texttt{zero}.

In addition we added 10 domain specific relations: \texttt{height}, \texttt{population}, \texttt{size}, \texttt{elevation}, \texttt{flights}, \texttt{objects}, \texttt{price}, \texttt{date}, \texttt{true}, \texttt{false}.

\subsection{Neural Models Parameters}
\label{subsec:supplemental_neural}

The \seqtoseq{} and \copynet{} models share similar encoder-decoder network architecture, as implemented in the AllenNLP library\footnote{\url{https://allennlp.org/}}. Parameters and hyperparameters used for training each of the final models, following parameters tuning on the development set of \textsc{Break}, are provided in Table~\ref{tab:model_parameters}. 

\mg{We should complete this part after parameter tuning.}

\begin{table}[ht]
\begin{center}
\small{
\begin{tabular}{lcc}
\multicolumn{1}{c}{} & \seqtoseq{} & \copynet{} \\
\hline
encoder layers & 5 & 5 \\
hidden dimension & 300 & 300 \\
dropout rate & 0.1 & 0.2 \\
learning rate & 0.14 & 0.07 \\
beam width & 5 & 5 \\
\hline
\end{tabular}}
\end{center}
\caption{Parameters and hyperparameters of the \seqtoseq{} and \copynet{} models.}
\label{tab:model_parameters}
\end{table}

\subsection{\textit{GED+} Implementation Details}
\label{subsec:ged_plus}

\textit{GED+} is an extended version of \textit{GED}, which includes edit operations for node merging and splitting. 
Here we describe the cost functions implemented for these operations. The cost of basic edit operations of addition, deletion and substitution of nodes and edges was remain the same as described for \textit{GED} in Section~\ref{sec:question_decomposition}.

Let $V=\{v_1, v_2, ..., v_{|V|}\}$ be a set of nodes, and denote by $P(V)$ the set of all permutations of $V$. Also, for a permutation $p=\langle v_{i_1}, v_{i_2}, ..., v_{i_{|V|}} \rangle$, define $C(p)$ as the concatenation of the node texts, in the order $p$, namely, $C(p)=v_{i_1} \oplus ... \oplus v_{i_{|V|}}$.
The cost function for splitting a node $v$ into a set of nodes $V$ is defined by:
$$|V| \cdot \min_{p\in P(V)}{\{1-\textit{MatchRatio}(v, C(p))\}}$$
In words, we find the permutation of $V$ with maximal \textit{MatchRatio} $c$ to $v$, and penalize each node in $V$ with the cost $1-c$.
Calculation of the cost for merging a set of nodes $V$ into a node $v$ is symmetric.
    
In addition, given the edit path between $G_{\decomplist}= \langle V_{\decomplist}, E_{\decomplist} \rangle$ and $G_{\hat{\decomplist}} = \langle V_{\hat{\decomplist}}, E_{\hat{\decomplist}} \rangle$, we penalize edit operations that modify the node order. This is done by creating a graph $G=\langle V=V_{\decomplist}\cup V_{\hat{\decomplist}}, E \rangle$ where
$$E = \{ (u,v) \;|\; \substack{v\in V_{\decomplist} , u\in V_{\hat{\decomplist}} ,\\ \exists \text{ substitution op mapping between $u,v$}\}} $$

is the set of edges resulting by node substitution operations.
We charge a cost of $1$ for every pair of crossing edges in $G$. The set of all such pairs is calculated using the classic sweep-line algorithm \cite{fortune1987sweepline}.

\subsection{RC Model Implementation}
\label{subsec:hotpot_implementation}

\paragraph{\textsc{BERT}}
To finetune BERT (base-uncased) on \textsc{HotpotQA}, we begin by converting each sample into \textsc{SQuAD} format. Paragraphs are ordered by their TF-IDF similarity to the question. The paragraph title and tokens "yes", "no" are appended to each paragraph. We consider the answer span to be the first span to occur in the supporting facts. We then finetune using \textsc{Hugging Face} \footnote{\url{https://github.com/huggingface/pytorch-transformers}}. The resulting model achieves 67.3 $F_1$ on the \textsc{HotpotQA} distractor dev set. We trained for 3 epochs with a batch size of 49, learning rate 3e-5, maximum sequence length 500 and document stride of 128.

\paragraph{\textsc{BreakRC}}
We use the single-hop RC model trained over \textsc{SQuAD}, following\footnote{\url{https://github.com/shmsw25/DecompRC/tree/master/DecompRC}} \cite{min2019multi}. First \textsc{SQuAD} is converted to a multi-paragraph setting. Added to the original context are $N$ additional Wikipedia paragraphs using the TF-IDF retriever. Finally, 3 instances of BERT (base-uncased) are finetuned with $N$ $=$ 0, 2, 4 to form an ensemble model. 

As an illustrative example of \textsc{BreakRC}, consider the question \textit{"Which 1970's film was released first, Charley and the Angel or The Boatniks?"} in Figure \ref{figure:hotpotqa_decompositions}. The \datasetname{} decomposition of the question has three steps. References \#1, \#2 denote the execution results of the single-hop RC model on steps \#1 and \#2 of the decomposition. For step \#1 \textsc{BreakRC} retrieves 10 paragraphs from the IR system and sends them to the single-hop RC which returns "23 march 1973". Similarly, it returns "1 july 1970" for step \#2. \textsc{BreakRC} will recognize step \#3 as a \texttt{SUPERLATIVE} step over "23 march 1973" and "1 july 1970". Finally, it will retrieve the \texttt{argmin} result using a symbolic date comparison.

}
\end{document}